\documentclass[AMS,STIX1COL]{WileyNJD-v2}

\usepackage{subcaption}
\usepackage{stix}
\usepackage{amsmath,amssymb,amsfonts}

\newcommand{\tr}{\textup{trace}}
\newcommand{\norm}[1]{\left\Vert#1\right\Vert}
\newcommand{\R}{\mathbb{R}}
\newcommand{\mydash}{\hbox{\sout{ }}} 
\newcommand{\Dcal}{\mathcal{D}}

\DeclareMathOperator*{\argmax}{arg\,max}

\articletype{Research Article}%

\received{}
\revised{}
\accepted{}

\begin{document}

\title{Vertex Nomination via Seeded Graph Matching}

\author[1]{Heather G. Patsolic}

\author[2]{Youngser Park}

\author[3]{Vince Lyzinski}

\author[1,2]{Carey E. Priebe*}

\authormark{H.G. Patsolic \textsc{et al}}

\address[1]{\orgdiv{Dept. of Applied Mathematics and Statistics},
    \orgname{The Johns Hopkins University},
    \orgaddress{\state{Maryland}, \country{USA}}}

\address[2]{\orgdiv{Center for Imaging Science}, \orgname{The Johns
        Hopkins University}, \orgaddress{\state{Maryland}, \country{USA}}}

\address[3]{\orgdiv{Dept. of Mathematics},
    \orgname{University of Maryland, College Park},
    \orgaddress{\state{Maryland}, \country{USA}}}

\corres{*Carey E. Priebe,\\ Whitehead Hall, 3400 N Charles St.\\ Baltimore, MD 21218\\
    \email{cep@jhu.edu}}


\abstract[Summary]{
Consider two networks on overlapping, non-identical vertex sets.
Given vertices of interest in the first network, we seek to identify
the corresponding vertices, if any exist, in the second network.
While in moderately sized networks graph matching methods can be applied directly to
recover the missing correspondences, herein we present a principled
methodology appropriate for 
situations in which the networks are too large/noisy for brute-force graph matching.
Our methodology identifies vertices in a local neighborhood of the
vertices of interest in the first network that have verifiable
corresponding vertices in the second network.
Leveraging these known correspondences, referred to as seeds,
we match the induced 
subgraphs in each network generated by the neighborhoods of these
verified seeds, and rank the vertices of the second
network in terms of
the most likely matches to the original vertices of interest. 
We demonstrate the applicability of our methodology through
simulations and real data examples. 
}


\jnlcitation{\cname{%
\author{H.G. Patsolic}, 
\author{Y. Park}, 
\author{V. Lyzinski}, and
\author{C.E. Priebe}} (\cyear{2019}), 
\ctitle{Vertex Nomination via Seeded Graph Matching},
\cjournal{Statistical Analysis and Data Science}}

\maketitle

\footnotetext{\textbf{Abbreviations:} VN, vertex nomination; SGM, seeded
    graph matching; VOI, vertex of interest}

\section{Introduction and Background}
\label{S:intro}

In this paper, we address the problem of nominating vertices across a pair of networks:  Given vertices of interest (VOIs) in a network $G=(V,E)$, our task is to identify the corresponding vertices of interest, if they exist, in a second network $G'=(V',E')$.  
Our methods will leverage vertices in the neighborhood of the VOIs in $G$ that have verifiable matches in $G'$ to (ideally) create local neighborhoods of the VOIs in both $G$ and $G'$.  
These neighborhoods are then soft-matched (see Algorithm \ref{alg:softsgm}, adapted here from \cite{fiadpr12}) across networks, yielding a nomination list for each VOI in $G$; i.e., a ranking of the vertices in the local neighborhood of the seeds in $G'$, ideally with the corresponding VOI's in $G'$ concentrating at the top of the list.
While global methods can (and have been) applied to identify the VOI's in $G'$ directly, performance of these methods can suffer from the noise induced by vertices without correspondences across networks \cite{lica15}.
    Localization is a prominent tool used across various fields such
    as machine learning (see for example \cite{priebe2005scan,wang2013locality}
on using locality based anomaly detection in time series of graphs and
\cite{GonenLMK} on localized multiple kernel learning),
pattern recognition (this includes clustering
algorithms which have been using localization for many years --- for
example $k$-nearest neighbor based classification rules --- see for example 
\cite{stone77,fukunage1975branch,keller1985fuzzy,muja2009fast,dgl}), and
object recognition (see for example \cite{sermanet2013overfeat} on using
convolutional networks for localization and object boundary detection
and \cite{beveridge1994local} on local algorithms for geometric object recognition).
Inspired by the many successes localization has seen in other fields of research,
we bring the concept of localization to the fore-front of network alignment.
Our methods are inherently local, leveraging recent advancements in both graph
matching \cite{fiadpr12,lyfifivoprsa15} and vertex nomination
\cite{coppersmith2012vertex,filypachpr15} to nominate across essentially
arbitrarily large networks.

Formally, suppose we are given two networks $G=(V,E)$ and $G'=(V',E')$ on overlapping but not necessarily identical vertex sets $V$ and $V'$ respectively.
For simplicity, we will presently restrict our attention to the case of a single VOI in $G$ (as the case of multiple VOIs is an immediate extension of our methodology for a single VOI), and
we write 
$$V=\{x\}\cup S \cup W \cup J,\,\,\,\,\,\, V'=\{x'\} \cup S' \cup W' \cup J',$$ where 
$x$ and $x'$ represent the VOI in $G$ and $G'$ resp.;
$S$ and $S'$ represent the seeded vertices across networks---those vertices that appear in both vertex sets whose correspondence across networks (i.e., the seeding $S\leftrightarrow S'$) is known a priori---and necessarily satisfy $s:=|S|=|S'|$;  
$W$ and $W'$ are the shared non-seed vertices---those vertices that appear in both vertex sets whose correspondence across networks is unknown a priori---with $|W| = |W'| =
n$; and $J$ and $J'$ are the unshared vertices---those vertices that appear in only one or the other vertex set without correspondences across networks---with $|J| = m$ and 
$|J'| = m'$.
Thus, we can write
$$\eta:=|V|=1+s+n+m,\,\,\, \text{ and } \eta'=|V'|= 1 + s + n + m'.$$ 
While the correspondence between vertices in $W$ and $W'$ is unknown a
priori, we will further assume that it is unknown which vertices in $G\setminus\{x,S\}$ are in $W$ versus $J$, as are the values of $n,m$ and $m'$.
Our inference task is to identify $x'\in V'\setminus S'$ (i.e., the corresponding VOI in $G'$) using only the knowledge of the graph structures and the
correspondence $S \leftrightarrow S'$.
For the purposes of this paper, we will
assume that the corresponding vertex $x'$ \textit{does} exist in $G'$, else our task is impossible.
Our goal will be to nominate vertices in $G'$ in a principled
manner so that the true match is high in the nomination list, thus
saving the end-user time in searching for this true match.
While this core-junk network framework has appeared often in the literature (see, for example, \cite{kazemi2015can}), herein we will consider a more general random graph model that allows for heterogeneity in vertex degree and behavior (see, Section \ref{sec:rhocorr}).

Our approach to this inference task lies on the boundary between 
{\it Graph Matching} and {\it Vertex Nomination}.
Stated simply, the formulation of the graph matching problem (GMP)
considered herein seeks to align the vertices in two networks so as
to minimize the number of induced edge disagreements between the aligned
networks.
Graph matching has been been extensively studied in the literature (for an excellent survey of the 
literature, see \cite{cofosave04, fopeve14}) with applications across various fields
including 
pattern recognition 
(see, for example, \cite{bebema05,wifekrma97,zhou2012factorized}),
computer vision (see, for example, \cite{zassvision1,leorVision1,yanVision1}), 
and biology (see, for example, 
\cite{zaslavskiy2009global,milenkovic2010optimal,kuchaiev2011integrative}), 
among others.
The seeded graph matching \texttt{SGM} algorithm on which
we base our primary algorithm has 
run-time $O(n^3)$ at worst, which
has been shown to be reasonable in comparison to other state-of-the-art
algorithms (such as the PATH algorithm of \cite{zaslavskiy2009path} --
see \cite{jovo_etal15} and \cite{fiadpr12} for more information 
on the computational complexity of this algorithm).
Furthermore, the authors of \cite{lyfipr14,lyfifivoprsa15} show that it
has theoretical guarantees for converging
to the correct solution under reasonable model assumptions.

The classical formulation of the vertex nomination (VN) inference task
\cite{marchette2011vertex,coppersmith2014vertex,sun2012comparison,suwan2015bayesian,filypachpr15,lyzinski2016consistency} 
can be stated as follows: given a network with latent community structure 
in which one of the communities is of particular interest and
given a few vertices from the community of interest, 
the task in vertex nomination is to order the remaining vertices in the 
network into a nomination list, with the aim of having vertices from the 
community of interest concentrate at the top of the list.
Thus, vertex nomination can also be thought of as a method for inferring
missing vertex labels, and is related to the class/labeled instances
acquisition task and collective classification methods of
\cite{bergsma2013using,talukdar2010,sen2008collective}.
The goal of vertex nomination is similar in spirit to popular network-based
information retrieval (IR) procedures such as \texttt{PageRank}
\cite{page1999pagerank} and personalized recommender systems on graphs
\cite{huang2004graph}.  
However, this formulation of VN is distinguished from other
supervised network mining tasks in both the generality of what defines a vertex
of interest \cite{rastogi2017vertex,lyzinski2017consistent}
and the (often) limited nature of the available training data (i.e., known
vertices of interest) in $G$.
Our present task can be viewed as vertex nomination
{\it across} networks: for a vertex of interest in $G$, we use graph matching
methodologies to order the vertices in $G'$ into a nomination list,  with the
aim of having the corresponding vertex of interest in $G'$ near the top of the
list. 

\paragraph{Our contributions:} In summary, our contributions are as follows: 
\begin{itemize}
    \item Leveraging the idea of principled sub-sampling of a graph, we
        reduce time-complexity for matching two graphs via
        \textit{localization}.
    \item 
            Combining the task of vertex nomination to across graph
            nomination tasks.
    \item 
        Extending the \texttt{SoftSGM} algorithm of \cite{fiadpr12} to
        the task of vertex nomination.
    \item Demonstrating via two real world graph data-sets, we conduct
        an out-of-the-box large-scale evaluation of our \texttt{VNmatch}
        algorithm.
\end{itemize}

The remainder of the paper is laid out as follows: 
In Section \ref{sec:relwork}, we give an overview of some related
work, after which, 
in Section \ref{sec:rhocorr}, we introduce a formal definition of what
we mean by "corresponding vertices."
Following, 
in Section \ref{sec:vnscheme}, we introduce our across graph VN scheme,
\texttt{VNmatch}, along with a brief mathematical description of the
utilized subroutines including
the soft seeded graph matching algorithm (\texttt{SoftSGM}, Algorithm \ref{alg:softsgm}), introduced in \cite{fiadpr12}.
In 
Sections \ref{sec:sbmexp} and \ref{sec:realdata}, we explore applications of \texttt{VNmatch} on both synthetic and real data, including a pair of high school friendship networks and a pair of online social networks.
We conclude with an overview of our findings and a discussion of potential extensions in Section \ref{sec:conc}.

We employ the \texttt{SoftSGM} Algorithm of \cite{fiadpr12}
as a means by which to nominate vertices in the \texttt{VNmatch}
algorithm so as to introduce this algorithm as a useful tool in the
across graph vertex nomination task. 
However, other vertex nomination schemes exist
which could also be adapted and utilized in the \texttt{VNmatch}
algorithm (in particular Steps 3 and 4).
%
%
For example, the use of spectral methods, which tend
to work well for matching larger graphs, may be desirable when the
original graphs are on the order of millions of vertices and
localization trims the networks down to only thousands of vertices.
For details regarding adjacency or Laplacian spectral embedding,
see \cite{twotruths}.


\paragraph{Notation:} To aid the reader, 
we have collected the frequently used notation introduced
in this manuscript into a table for ease of reference; see Table 
\ref{table:notation}\hspace{-5pt}.
Also, in what follows, we assume for simplicity that all graphs are simple (that
is, edges are undirected, there are no multi-edges, and there are no
loops).

\begin{table*}[ht!]
    \centering
\caption{ Table of frequently used notation}
  \label{table:notation}
\begin{tabular}{|l|l|}
\toprule
    Symbol & Description\\ 
\midrule
    $G=(V,E)$ & A graph with vertex set, $V$, and edge set, $E$ \\
    $G[T]$ & For $T\subset V$, this is the induced subgraph of $G$ on $T$ \\
    $\eta$ (resp., $\eta'$) & Number of vertices in $V$ (resp., $V'$)\\ 
    $S$ (resp., $S'$) & Set of $s$ seed vertices in $G$ (resp., $G')$\\ 
    $x$ (resp., $x'$)& Vertex of interest in $G$ (resp., $G'$)\\ 
    $W$ (resp., $W'$) & Set of all $n$ shared non-seed and non-VOI vertices in $G$ (resp., $G'$)\\ 
    $U$ (resp., $U'$)& Set of all shared vertices, including seeds and VOI in $G$ (resp., $G'$)\\ 
    $J$ (resp., $J'$)& Set of $m$ (resp., $m'$) unshared vertices in $G$ (resp., $G'$)\\ 
    $H$ (resp., $H'$)& $G[U]$ (resp., $G'[U']$)\\ 
    $N_t(T)$, & Set of all vertices within $t$-length path of
    $T\subset V $ (including $T$) \\ 
    $I_n (0_n)$ & The $n\times n$ identity (zeroes) matrix \\
    $\mathbb{1}$ & Appropriately sized vector of all ones \\
    $\Pi_k$ & Set of all $k\times k$ permutation matrices \\
    $\Dcal_k$ & Set of all $k\times k$ doubly stochastic matrices \\
    $h$ & Maximum considered path length from seeds to VOI, $x \in
    G$ \\
    $S_x$ and $S_x'$ & $S_x=S\cap N_h(x)$ with corresponding seeds $S_x'$ in $V'$\\
    $\ell$ & Maximum path length for neighborhood around $S_x$ \\
    $G_x=(V_x,E_x)$ (resp., $G_x'=(V_x',E_x')$) & $G[N_\ell(S_x)]$ (resp., $G'[N_\ell(S_x')]$)\\
    $C_x'$ & The set of candidate matches for $x$ in $V_x'$, namely $C_x'=V_x'\setminus S_x'$\\
    $\Phi_x$ & Nomination list output from Algorithm \ref{alg:vnsgm} \\
    $\tau(x')$ & Normalized expected location of $x'$ in $\Phi_x$\\
    $|T|$ & Cardinality of set $T$\\
    $\norm{A}_F$ & Frobenius norm of matrix $A$\\
    $\oplus$ & direct sum \\
\bottomrule
\end{tabular}
 \end{table*}    

\section{Related Work}
\label{sec:relwork}

A number of inexact graph matching algorithms have been
extended/developed recently to match
graphs with overlapping, non-identical vertex sets. 
Two such algorithms include percolation based algorithms (see for
example \cite{kazemi2015can,fabiana2015anonymizing,kahagr15}) 
and Bayesian based algorithms (see for example \cite{pedarsani2013bayesian}).

In \cite{kazemi2015can} and \cite{kahagr15}, the authors focus their
efforts on proving that under the independent-edge-sampling model 
$G(n,p;t.s)$, where
a graph, $G$, is generated from an Erd\"os-Renyi $(n,p)$ model and two
subgraphs of $G$, namely $G_1$ and $G_2$, are generated so that the
probability of a node from $G$ belonging to $G_i$ is $s$,
independently for $i=1,2$, and similarly for edges (with
probability $t$).
Under the independent-edge-sampling model, 
it is shown in \cite{kazemi2015can} that for sufficiently large $p$ the true
partial matching is recoverable under particular model assumptions and
for some formulation of their objective; however, as the authors admit,
the optimization formulation proposed is not scalable, and there is no
mention of how the correct formulation of the objective is to be
obtained. 

Using the same independent-edge-sampling model, the authors of
\cite{kahagr15} introduce an iterative percolation based graph matching
method for seed-based graph matching, demonstrating that their method
(under this model and particular assumptions) matches nearly all
overlapping nodes correctly.  
In \cite{fabiana2015anonymizing}, the authors introduce a degree-driven
percolation based graph matching algorithm which uses an iterative
approach to match nodes with higher degree to lower degree using
percolation based graph matching. 
For scale-free networks, the authors
show that, under particular model assumptions,
their method, which does not aim to match all nodes, but to
match subsets of nodes from the two graphs, 
matches nearly all vertices which have a match correctly and 
that the algorithm does not match any nodes incorrectly.
While this works well for scale-free networks, the advantages of this
method would be more limited on graphs with more block structure and without
having higher-degree nodes which help seed the rest of the algorithm.
The authors point out that when seeds are chosen uniformly at random 
$o(n^{1/2+\epsilon})$ seeds are needed to match most vertices correctly,
but allowing for more intelligent seed-selection based on vertex degree,
as few as $n^{\epsilon}$ seeds may be sufficient, for some arbitrarily
small $\epsilon$.

Each of the above approaches is theoretically based in relatively simple random graph models (ER for \cite{kazemi2015can} and the Chung-Lu model in \cite{fabiana2015anonymizing}), while also demonstrating good performance in more complex real data settings.
Our present approach is naturally situated in the more general Random Dot Product Graph setting of \cite{young2007random}.  
While still not able to capture all the intricacies of real network data, the random dot product graph is quite flexible and encompasses numerous other common random graph models (ER, Chung-Lu, positive definite stochastic blockmodel, etc.).
In addition, we also demonstrate the effectiveness of our method on more complex real data networks as well.

Percolation based algorithms could certainly be used for vertex
nomination in a similar way that we present vertex nomination based on
the seeded graph matching algorithm of \cite{fiadpr12} 
(which is based on a fast approximate quadratic programming algorithm 
of \cite{jovo_etal15}). 
One of the advantages of the present optimization based approach is the ability to efficiently explore the space of locally optimal solutions near the global optimum.
Practically, the graphs to be matched
in real data are much more messy than theory would allow, and the variations
that can be obtained from these local optima provide a degree of robustness to model misspecification.
Furthermore, the SGM algorithm itself
runs quickly on modestly sized networks and has asymptotic guarantees
for particular models and conditions 
(see for example \cite{jovo_etal15,fiadpr12} and \cite{lyfipr14,lyfifivoprsa15}).

In \cite{ji2016relative} the authors focus on the task of
de-anonymizability, and
explore a method for matching nodes based on node-degree; that
is, the authors consider two graphs drawn in some manner from a larger
graph and attempt to de-anonymize (match) the vertices in the two graphs 
which have the highest degrees. 
We are not concerned with matching vertices based on their degree, since
a vertex of interest is based on an external characteristic that is not
necessarily related to the degree distribution of the two graphs. 

Another technique for approximate graph matching relies on Bayesian
methods \cite{pedarsani2013bayesian}. 
The authors of \cite{pedarsani2013bayesian}
introduce a method which relies on estimating the posterior probability
that two nodes should be matched based on a particular prior. In the
afore-mentioned paper, the authors rely on node attributes, such
as vertex degree, mapping a few nodes at a time in an iterative manner
until all nodes are matched; any nodes matched in one iteration will be
used as seeds (referred to as \textit{anchors}) in the next iteration. 
In the end, the authors seek to obtain a hard matching of the nodes that
maximizes the sum of the log-posteriors for all node pairs.
While the idea of a posterior probability that two nodes
should be matched is a similar idea to what we present, the purpose of
our more frequentist method is to utilize a \textit{soft matching} of
the nodes in order to rank them in order from most to least likely
matches to the vertex (or vertices) of interest. 

\section{Corresponding vertices}
\label{sec:rhocorr}

Consider two social networks 
in which vertices represent users/accounts and edges represent whether
or not two accounts are linked in some way.
An individual may have an account on one network or the other or both.
We would say that two accounts across the platforms correspond to each
other if the same individual runs both accounts;
that is, both nodes correspond to the same individual although with possibly different node labels.
Arguably an individual who has an account on both networks will have
similar, though not identical, behavior 
across the two networks.  
Consider an email network in which nodes are email
addresses and two email addresses share 
an edge (directed or not) if they send correspondence to one another,
and a phone network in which nodes are phone numbers and
edges represent whether or not one of the numbers calls the other.
In this example, a vertex in the email network will correspond to a
vertex in the call network if the email and phone number belong to the
same individual.
An individual who uses both email and phone correspondence
may communicate with individuals in the two networks in a similar,
though not identical, manner.
Thus, if there
is a connection between two individuals in one network and those same
individuals exist in the second network, one would think that it is
more likely that there exists a connection between these individuals 
in the second network, i.e. 
there is a positive correlation between the
edges between these vertices across the networks. 
To model this correspondence, we proceed as follows.

With notation as above, let $U = \{x\} \cup S \cup W$ and $U' = \{x'\} \cup S' \cup W'$ denote
the set of shared vertices between $G$ and $G'$ with
$|U| = |U'| = 1 + s + n$.
Define $H:=G[U]$ (resp., $H'=G'[U']$), the induced subgraph of $G$ (resp., $G'$) on the vertex set $U$ (resp., $U'$). 
As $H$ and $H'$ are graphs on the same (though potentially differently labeled) vertex sets, 
we model a shared structure present across $H$ and $H'$ as $(H,H') \sim \rho\mydash 
RDPG(X)$.  Before defining the $\rho\mydash RDPG$ model, we first recall 
the definition of a {\it random dot product graph} (RDPG); 
see \cite{young2007random}.
\begin{definition}
\label{def:RDPG}
Consider $X = [X_1, \ldots, X_n]^\top \in \R^{n\times d}$ satisfying $XX^\top \in [0,1]^{n\times n}$.  
We say that graph $G$ with adjacency matrix $A$ is distributed as a random dot product graph with parameter $X$ (abbreviated $G \sim RDPG(X)$) if given $X$, 
\begin{equation*}
    A_{i,j} \stackrel{\text{ind.}}{\sim} Bernoulli(X_i^\top X_j),
\end{equation*} i.e., 
\begin{equation*}
    P(A|X) = \prod_{i<j} (X_i^\top X_j)^{A_{i,j}}(1-X_i^\top
    X_j)^{1-A_{i,j}}.
\end{equation*}
Conditioning on $X$, this is an independent-edge random
graph model where
vertex $v_i$ is associated with a latent position vector
$X_i\in\mathbb{R}^d$, and the probability of an edge between any two
vertices is determined by the dot product of their associated latent
position vectors.
\end{definition}


To imbue multiple random dot product graphs with a notion of vertex
correspondence, we 
correlate the behavior of nodes across networks.
We call this new model the $\rho\mydash RDPG$
model, which is defined as follows.

\begin{definition}
    \label{defn:rhordpg}
    Consider $X = [X_1, \ldots, X_n]^\top \in \R^{n\times d}$ satisfying $XX^\top \in [0,1]^{n\times n}$.  
The bivariate graph valued random variables $(G,G')$---with respective adjacency matrices $A$ and $A'$---are said to be distributed as a pair of $\rho$-correlated random dot product graphs with parameter $X$ (abbreviated $(G,G') \sim \rho\mydash RDPG(X)$) if 
    \begin{enumerate}
        \item Marginally, $G, G' \sim RDPG(X)$, and
        \item $\{A_{i,j},A'_{k,l}\}_{\{i,j\},\{k,l\}\in\binom{V}{2}}$ are collectively independent except that for each $\{i,j\}\in\binom{V}{2}$, 
        $\mathrm{correlation}(A_{i,j},A'_{i,j})=\rho.$
    \end{enumerate}
\end{definition}


Our framework posits $(H,H') \sim \rho\mydash 
RDPG(X)$ for a latent position matrix $X\in\mathbb{R}^{(1+s+n) \times d}$. 
In order to generate the full graphs $G$ and $G'$ which also have
unshared vertices, 
we generate $G\sim RDPG([X,Y])$ and $G'\sim RDPG([X,Y'])$,
so that the induced subgraphs $(H,H') \sim \rho\mydash RDPG(X)$ 
and the remaining edges of $G$ and $G'$ are formed independently as in
the case for the general RDPG. Thus, the first $1+s+n$ vertices in the
two graphs correspond to one another via the identity map and the
remaining $m$ and $m'$ vertices of $G$ and $G'$, respectively, represent
the unshared vertices $J$ and $J'$.
Here, $Y\in\mathbb{R}^{m \times d}$ and $Y'\in\mathbb{R}^{m' \times d}$ represent the respective latent positions for the unshared vertices in $G$ and $G'$.
For ease of notation, we will write 
$(G,G') \sim \rho\mydash RDPG(X,Y,Y')$, where $(G,G')$ 
is realized as two graphs:
$G$ on $\eta=1+s+n+m$ vertices $\{x\} \cup S \cup W \cup J$ and 
$G'$ on $\eta'=1+s+n+m'$ vertices $\{x'\} \cup S' \cup W' \cup J'$.

If $G$ and $G'$ exhibit latent community structure, 
it can be fruitful to model them as Stochastic block model 
(SBM) random graphs \cite{Holland1983}. 
SBMs have been extensively studied in the literature and have been shown
to provide a useful and theoretically tractable model for more complex
graphs with underlying community structure
\cite{rohe2011spectral,STFP,Airoldi2008,olhede_wolfe_histogram}.
We define the stochastic block model as follows.
\begin{definition}
    \label{def:sbm}
    We say that graph $G=(V,E)$ with adjacency matrix $A$ is distributed as a stochastic block model random graph with parameters $k$, $b$ and $\Lambda$ (abbreviated $G\sim\text{SBM}(k,b,\Lambda)$) if 
    \begin{enumerate}
        \item $V$ is partitioned into $k$ blocks $V=V_1\cup V_2\cup\cdots\cup V_k$;
        \item $b: V \to \{1,\ldots,k\}$ is a map such that $b(i)$
            denotes the block label of the $i^{\text{th}}$ vertex;
        \item $\Lambda \in [0,1]^{k\times k}$ is a matrix such that 
            $A_{i,j}\stackrel{ind}{\sim}$Bernoulli$(\Lambda_{b(i),b(j)})$ 
            for distinct $\{i,j\} \in \binom{V}{2}$.
    \end{enumerate}
\end{definition}
\noindent Recall that the edge probability matrix for a random dot product graph
model, with parameter $X\in \R^{n\times d}$, is equal to $XX^T$, which is positive
semi-definite. If $X$ consists of precisely $k$ distinct rows, then 
a graph generated via $RDPG(X)$ can also be said to be generated from 
a stochastic block model having
$k$ blocks, block assignment vector $b$ assigning vertices with the same
latent position to the same block, and probability matrix $\Lambda = X^{(k)}(X^{(k)})^T$
where $X^{(k)}$ here refers to the $k\times d$ matrix of distinct rows of $X$.
Moreover, if $\Lambda$ is positive semidefinite, then
$G\sim\text{SBM}(k,b,\Lambda)$ can be realized as a RDPG with
appropriately defined $X$.
Thus, there is an overlap in the set of random dot product graph models
and stochastic block models.

We can then define the $\rho\mydash SBM$ model as follows. 
\begin{definition}
    \label{def:rhosbm}
   The bivariate graph valued random variables $(G,G')$---with
   respective adjacency matrices $A$ and $A'$---are said to be
   distributed as a pair of $\rho$-correlated stochastic block model
   graphs with parameter $k$, $b$, and $\Lambda$ (abbreviated $(G,G')
   \sim \rho\mydash \text{SBM}(k,b,\Lambda)$) if 
    \begin{enumerate}
        \item Marginally, $G, G' \sim \text{SBM}(k,b,\Lambda)$, and
        \item $\{A_{i,j},A'_{k,l}\}_{\{i,j\},\{k,l\}\in\binom{V}{2}}$ are collectively independent except that for each $\{i,j\}\in\binom{V}{2}$, 
        $$\text{correlation}(A_{i,j},A'_{i,j})=\rho.$$
    \end{enumerate}
\end{definition}
\noindent Note that if we generate $H$ and $H'$ from a 
$\rho\mydash SBM(k,b,\Lambda)$, $G$ and $G'$ can be constructed so
that
$$G\sim \rho \mydash SBM(k_1,b_1,\Lambda_1)\text{ and }G'\sim \rho \mydash SBM(k_2,b_2,\Lambda_2),$$ where
$k \leq \min(k_1,k_2),$ $b_1(j) = b_2(j)= b(j)$ for all $j \in \{1,2,\ldots, 1+s+n\}$,
and the upper left $k\times k$ submatrix of $\Lambda_i$ is $\Lambda$
(for $i=1,2$). 
We write this formally as 
$$(G,G')\sim \rho \mydash SBM(k_1,k_2,b_1,b_2,\Lambda_1,\Lambda_2).$$

\section{Vertex nomination via seeded graph matching}
\label{sec:vnscheme}

With this notion of corresponding vertices,
we next introduce our proposed algorithm for finding the corresponding
vertex $x' \in V'$ to a particular vertex of interest $x \in V$. 
Again, we assume a single vertex of interest for simplicity, as the
extension to multiple vertices of interest follows immediately. 
Before presenting our main algorithm, 
\texttt{VNmatch} (Algorithm \ref{alg:vnsgm}), we first provide the
necessary details for the subroutine of Algorithm \ref{alg:vnsgm} we
employ, namely the \texttt{SoftSGM} algorithm of \cite{fiadpr12}.
The easy interpretability and simple extension of the \texttt{SoftSGM}
algorithm to generating nomination
lists for vertices of interest make it a natural candidate for the
vertex nomination subroutine task of Algorithm \ref{alg:vnsgm}; however,
other methods of graph matching, such as spectral-based methods,
for which extension to vertex nomination
is possible could also be used during this step of the algorithm.

\subsection{Soft seeded graph matching}
\label{sec:sgmintro}

Given $A$ and $A'$ in $\mathbb{R}^{n\times n}$, the respective adjacency matrices of two $n$-vertex graphs $G$ and $G'$, the {\it graph matching problem} (GMP) is
\begin{equation}
\label{eq:GMP}
\min_{P\in\Pi_n}\|AP-PA'\|_F,
\end{equation}
where $\Pi_n$ is the set of $n\times n$ permutation matrices,
and $\norm{M}_F$ denotes the Frobenius norm of the matrix $M$. 
While the formulation in Eq.\@ (\ref{eq:GMP}) seems restrictive, it is easily adapted to handle the case where the graphs are weighted, directed, loopy and on potentially different sized vertex sets (using, for example, the padding methods introduced in \cite{fiadpr12}).

In our present setting, where we have 
known seeded vertices $S\leftrightarrow S'$, we consider the closely
related {\it seeded graph matching problem} (SGMP) (see, for example,
\cite{fiadpr12,lyfipr14,lyfifivoprsa15, 
   ham2005semisupervised,fopeve14,lica15,lyzinski2015spectral}).
We have $G=(V,E)$ and $G'=(V',E')$ with vertex sets 
$V = \{x\}\cup S\cup W \cup J$ and $V' = \{x'\} \cup S' \cup W' \cup J'$, 
with $|V| = \eta = 1 + s + n + m$, 
$|V'| = \eta' = 1+ s+n + m' $,
and seeding $S \leftrightarrow S'$.
Without loss of generality, suppose $S = S' = \{1,2,\ldots,s\}$ (if no seeds are
used, $s=0$ and $S = S' = \emptyset$), and suppose for the moment that 
$\eta= \eta'$.
The seeded graph matching problem aims to solve 
\begin{equation}
    \label{eqn:sgmobj}
    \min_{P \in \Pi_{\eta-s}}
    \norm{A(I_s \oplus P) - (I_s\oplus P)A'}_{F}^2, 
\end{equation}
where, $\Pi_{\eta-s}$ denotes the set of $(\eta-s) \times (\eta-s)$ permutation
matrices, and $\oplus$ denotes the direct sum of matrices.
Note that decomposing $A$ and $A'$ via
\begin{equation}
    \label{eqn:ABblock}
A= \begin{pmatrix} A_{11}& A_{12} \\ A_{21} & A_{22} \end{pmatrix}, \text{ and }
A'=\begin{pmatrix} A'_{11}& A'_{12} \\ A'_{21} & A'_{22} \end{pmatrix} 
\end{equation}
where $A_{11},\,A'_{11} \in \mathbb{R}^{s\times s}$, 
$A_{21}^T,\,(A'_{21})^T,A_{12},\,A'_{12}\in \mathbb{R}^{(\eta-s)\times s}$, 
and $A_{22},\,A'_{22}\in \mathbb{R}^{(\eta-s)\times (\eta-s)}$, 
the SGMP is equivalent to 
\begin{multline}
\label{eq:traceP}
\max_{P\in \Pi_{\eta-s}} f(P) = 
\max_{P\in \Pi_{\eta-s}} 
\left[ \tr(P^T A_{21}(A'_{21})^T)+\right. \\
\left. \tr(P^TA^T_{12}A'_{12})+ 
\tr(A_{22}^TPA'_{22}P^T)\right].
\end{multline}

The SGMP, in general, is NP-hard, and  
many (seeded) graph matching algorithms begin by relaxing the feasible region of 
Eq. (\ref{eqn:sgmobj}) or (\ref{eq:traceP}) from the discrete
$\Pi_{\eta-s}$ to the convex hull of $\Pi_{\eta-s}$ 
\cite{zaslavskiy2009path,fispvomusa13,jovo_etal15,fiadpr12},
which, by the Birkoff-vonNeumann theorem, 
is the set of $(\eta-s)\times (\eta-s)$ 
doubly stochastic matrices, denoted $\Dcal_{\eta-s}$.
This relaxation enables the machinery of continuous optimization (gradient descent, ADMM, etc.) to be employed on the relaxed SGMP.
Note that while the solutions of Eq.~(\ref{eqn:sgmobj}) and 
(\ref{eq:traceP}) are equivalent, the solutions of the relaxations of 
Eq.~(\ref{eqn:sgmobj}) and (\ref{eq:traceP}) are not equivalent in general, with the indefinite relaxation, Eq. (\ref{eq:traceP}), preferable under the model assumptions we will consider in this paper \cite{lyfifivoprsa15}.

The \texttt{SGM} 
algorithm of \cite{fiadpr12} approximately solves this indefinite SGMP relaxation using the Frank-Wolfe algorithm \cite{frwo56}, 
and then projects the obtained doubly stochastic solution onto 
$\Pi_{\eta-s}$.
The algorithm performs excellently in practice in both synthetic and real data settings, with a $O((\eta-s)^3)$ runtime allowing for its efficient implementation on modestly sized networks. 
Since we ultimately aim to create a nomination list (and not a 1--to--1 correspondence necessarily) for the 
VOI of likely matches in $V'\backslash S'$, 
we use the \texttt{SoftSGM} algorithm of \cite{fiadpr12}---a stochastic averaging of the original \texttt{SGM} procedure over multiple random restarts---in order to softly match the graphs.
Rather than the 1--to--1 correspondence output from \texttt{SGM}, \texttt{SoftSGM} (pseudocode provided in Algorithm \ref{alg:softsgm} for completeness)
outputs a function $p(\cdot,\cdot):V\times V'\mapsto[0,1]$, where $p(i,j)$ represents the
likelihood vertex $j$ in $G'$ matches to vertex $i$ in $G$.   
As noted in Table \ref{table:notation}\hspace{-5pt}, $A\oplus B$ denotes the direct sum
between two matrices $A$ and $B$, and $0_{n}$ denotes the $n\times n$
all zeroes matrix.
Also, the function $f$ in Algorithm \ref{alg:softsgm} refers
to $f$ as in Equation \ref{eq:traceP}.

\begin{algorithm}[t!]
\begin{algorithmic}
\State \textbf{Input}:
$G\in\mathcal{G}_\eta, G'\in\mathcal{G}_{\eta'}$ with respective
adjacency matrices $A$ and $A'$; number of seeds $s$ (assumed to be
first $s$ vertices of $G$ and $G'$); number of random restarts
$R\in\mathbb{N};$  random initialization parameter
$\gamma\in[0,1];$ stopping criterion $\epsilon$;
\State \textbf{Step 0}: if $\eta \ne \eta'$ set $A = (2A - (\mathbb{1}\mathbb{1}^T - I_{\eta})) \oplus
0_{\min(0,\eta'-\eta)}$ and 
$A' = (2A' - (\mathbb{1}\mathbb{1}^T - I_{\eta'})) \oplus
0_{\min(0,\eta-\eta')}$;

\For{i=1:R}
\State \textbf{Step 1}:
Generate $Q_i$ Uniformly from the set of permutation matrices, $\Pi_{n-s}$;
\State \textbf{Step 2}:
Generate $\beta_i$ Uniformly from $(0,\gamma)$ and set 
$P^{(0)}_i=\beta_i Q_i+(1-\beta_i)\frac{1}{n-s} (\mathbb{1}\mathbb{1}^T);$
\State \textbf{Step 3}:
\While{$\|f(P^{(j)})-f(P^{(j-1)})\|_F>\epsilon$}
\State \textbf{Step a}:
Compute $\nabla f(P^{(j)}) = 
A_{21}B_{21}^T + A_{12}^T B_{12} +
A_{22}P^{(j)} B_{22}^T + A_{22}^T P^{(j)} B_{22}$; 

\State \textbf{Step b}:
Compute 
$Q^{(j)} \in \arg\max \{\tr (Q^T \nabla f(P^{(j)}))\}$ over 
$Q \in \Dcal_{n-s}$ via the Hungarian
Algorithm \cite{ku55}; 

\State \textbf{Step c}:
Compute 
$\alpha^{(j)} = \arg \max\{ 
    f(\alpha P^{(j)} + (1-\alpha)Q^{(j)})\}$ over $\alpha \in[0,1]$;

\State \textbf{Step d}:
Set $P^{(j+1)} = \alpha^{(j)} P^{(j)} + (1-\alpha^{(j)})
Q^{(j)})\}$;
\EndWhile
\State \textbf{Step 5}:
Compute $P_i \in \arg\max \{\tr(Q^T P^{(\text{final})})\}$ over $Q \in \Pi_{n-s}$ via the
Hungarian Algorithm, where $P^{(\text{final})}$ is output from the while
loop;
\EndFor
\State \textbf{Step 6}:
Define $p$ via $p(\ell,k) = \left[\sum_{i=1}^R \frac{1}{R} P_i\right]_{\ell,k}$;
\State \textbf{Output}: $p$
\end{algorithmic}
\caption{\texttt{SoftSGM}}
\label{alg:softsgm}
\end{algorithm}

\subsection{\texttt{VNmatch}}
\label{ssec:vnmatch}

We consider two graphs $G$ and $G'$ with vertex sets
$V = \{x\} \cup S \cup W \cup J$ and $V' = \{x'\} \cup S' \cup W' \cup
J'$, where the vertices in $V\backslash J$ and $V' \backslash J'$ are
shared between the two graphs.
As stated previously, our task is to 
leverage an
observed one-to-one correspondence $S\leftrightarrow S'$ 
to find the vertex $x' \in V'$ corresponding to a particular
vertex of interest $x \in V$.
If $G$ and $G'$ are modestly sized (on the order of thousands of
vertices), we could use Algorithm \ref{alg:softsgm}, the
\texttt{SoftSGM} algorithm of \cite{fiadpr12}, to
soft match $G$ and $G'$, padding $V$ or $V'$ as necessary when $\eta \ne \eta'$.
As the purpose of matching the graphs in this inference task is to identify the vertex $x' \in
V'$; we create a ranked nomination list, which we denote by $\psi_x$, for $x$ by ordering the vertices in
$G'$ by decreasing value of
$p(x,\cdot)$: (with ties broken uniformly at random)
\begin{align*}
    \psi_x(1)\in&\argmax\limits_{r\in v'}p(x,r),\\
\psi_x(2)\in&\argmax_{r\in (v')\setminus \{\psi_x(1)\}}p(x,r),\\
 &\vdots\\
 \psi_x(\eta')\in&\argmax_{r\in (v')\setminus
     \{\psi_x(1),\ldots,\psi_x(\eta'-1)\}}p(x,r).
\end{align*}

\begin{algorithm}[t!]
\begin{algorithmic}
\State \textbf{Input}:
$G,G'$; 
$S,S'$ -- the seed sets, with $S\leftrightarrow S'$; 
$\ell \geq h$
\State \textbf{Step 1}: 
find $S_x=s\cap N_h(x)$, and matching vertices $S_x\leftrightarrow S'_x$
in $G'$, if $|S_x|=0$, stop;
\State \textbf{Step 2}:
create $G_x=G[N_\ell(S_x)]$ and $G_x'=G[N_\ell(S'_x)]$;
\State \textbf{Step 3}:
use graph matching (we use \texttt{SoftSGM} Algorithm \ref{alg:softsgm}) 
to match $G_x=(V_x,E_x)$ and
$G'_x=(V_x',E_x')$, yielding $p(\cdot,\cdot):V_x\times V'_x\mapsto[0,1]$; 
\State \textbf{Step 4}:
create nomination list $\phi_x$ for $x$ by ranking the vertices in $V_x'$ by decreasing value of $p(x,\cdot)$;
\State{\bf Output}: $\phi_x$
\end{algorithmic}
\caption{\texttt{VNmatch}: vertex nomination via seeded graph matching}
\label{alg:vnsgm}
\end{algorithm}

In practice, however, the networks under consideration may be too large to directly apply \texttt{SoftSGM} or similar global graph matching procedures.
For example, many of the partially crawled social networks found at 
\cite{snapnets} contain tens-of-millions of vertices or more.
Therefore, rather than applying \texttt{SoftSGM} globally, 
we reduce the size of the problem through localization.
In our underlying network model,
the local structure around a
vertex in one graph will be similar to the local structure around a
vertex in the second graph. 
With this in mind, 
given $h \in \mathbb{N}$ and a set $\Upsilon \subset V$, 
we define the {\it h-neighborhood of $\Upsilon$} in $G$ via 
\begin{align*}
    N_h(\Upsilon) &:= \{ v\in V : \text{ there exists a path of length
        }\\
        & \hspace{19pt} \leq h
        \text{ in }G\text{ from } v \text{ to a vertex in } \Upsilon\}.
\end{align*}
Note, by convention $\Upsilon\subset N_h(\Upsilon)$.
We denote by $S_x =S_{x,h}:= S \cap N_h(x)$ the set of seeded vertices
in $G$ with shortest path distance to $x$ less than or equal to $h$, and
we define $S'_x$ to be the corresponding seeds in $G'$ with 
$|S_x| = s_x = |S'_x|$.
Notionally, as $h \rightarrow \infty$, $N_h(x)$ tends towards the
connected component of $G$ containing $x$,
and we say that $h = \infty$ yields $N_h(x)$ to be the entire 
vertex set $V$ of $G$.

For $\ell \geq h$, we define $G_x=(V_x,E_x):=G[N_{\ell}(S_x)]$ and 
$G'_x=(V_x',E_x'):=G[N_{\ell}(S'_x)]$
to be the respective induced subgraphs of $G$ and $G'$ generated by $N_{\ell}(S_x)$ and 
$N_{\ell}(S_x')$.
Ideally, $N_{\ell}(S'_x)$---which is a local $\ell$-neighborhood of those seeds in $G'$ whose distance to $x$ in $G$ is at most $h\leq \ell$---will contain $x'$, the corresponding VOI in $G'$.
If so, we propose to uncover the correspondence $x\leftrightarrow x'$ by using \texttt{SoftSGM} to soft match $G_x$ and $G_{x}'$ rather than all of $G$ and $G'$.
The output of \texttt{SoftSGM} is then $p(\cdot,\cdot):V_x\times
V_x'\mapsto[0,1]$, and 
we create the nomination list for $x$, denoted $\Phi_x$, by ranking the vertices in $V_x'$ based on decreasing value of $p(x,\cdot)$; i.e., if $|V_x'|=\xi$ then
\begin{align*}
    \Phi_x(1)\in&\argmax\limits_{r\in V_x'}p(x,r),\\
\Phi_x(2)\in&\argmax_{r\in (V_x')\setminus \{\Phi_x(1)\}}p(x,r),\\
 &\vdots\\
 \Phi_x(\xi)\in&\argmax_{r\in (V_x')\setminus
     \{\Phi_x(1),\ldots,\Phi_x(\xi-1)\}}p(x,r),
\end{align*} 
where ties are broken uniformly at random.

\begin{figure}[t!]
\centering
\begin{subfigure}[b]{0.4\textwidth}
    \includegraphics[width=\textwidth]{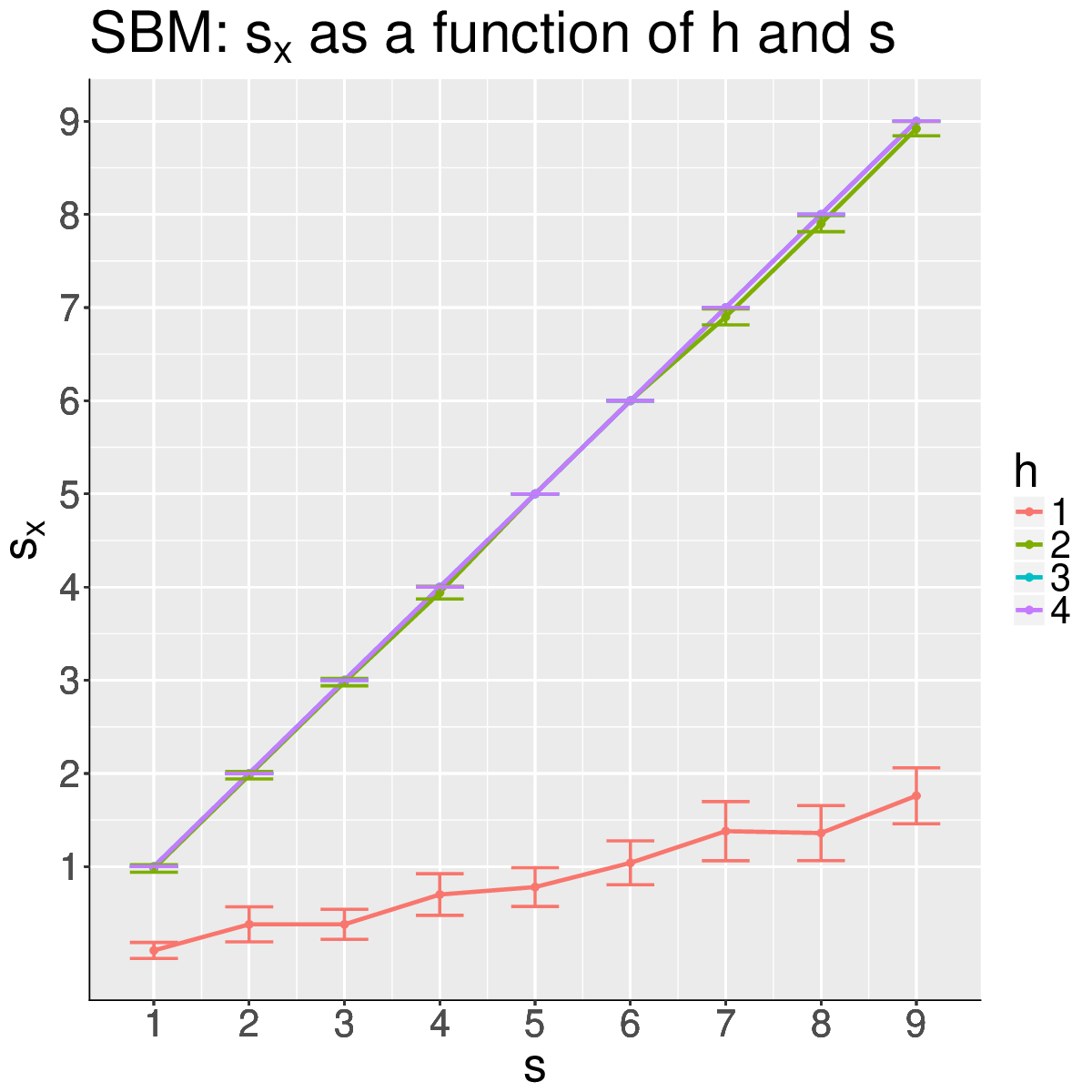}
    \caption{
            $SBM(3,b,\Lambda)$ on $300$ vertices, with eqaul block
        sizes. $\Lambda$ is defines so that the diagonal elements are
        $0.4$ and the off-diagonal elements are $0.05$.
    }
    \label{fig:sbm_funhs}
\end{subfigure}
\quad\quad
\begin{subfigure}[b]{0.4\textwidth}
    \includegraphics[width=\textwidth]{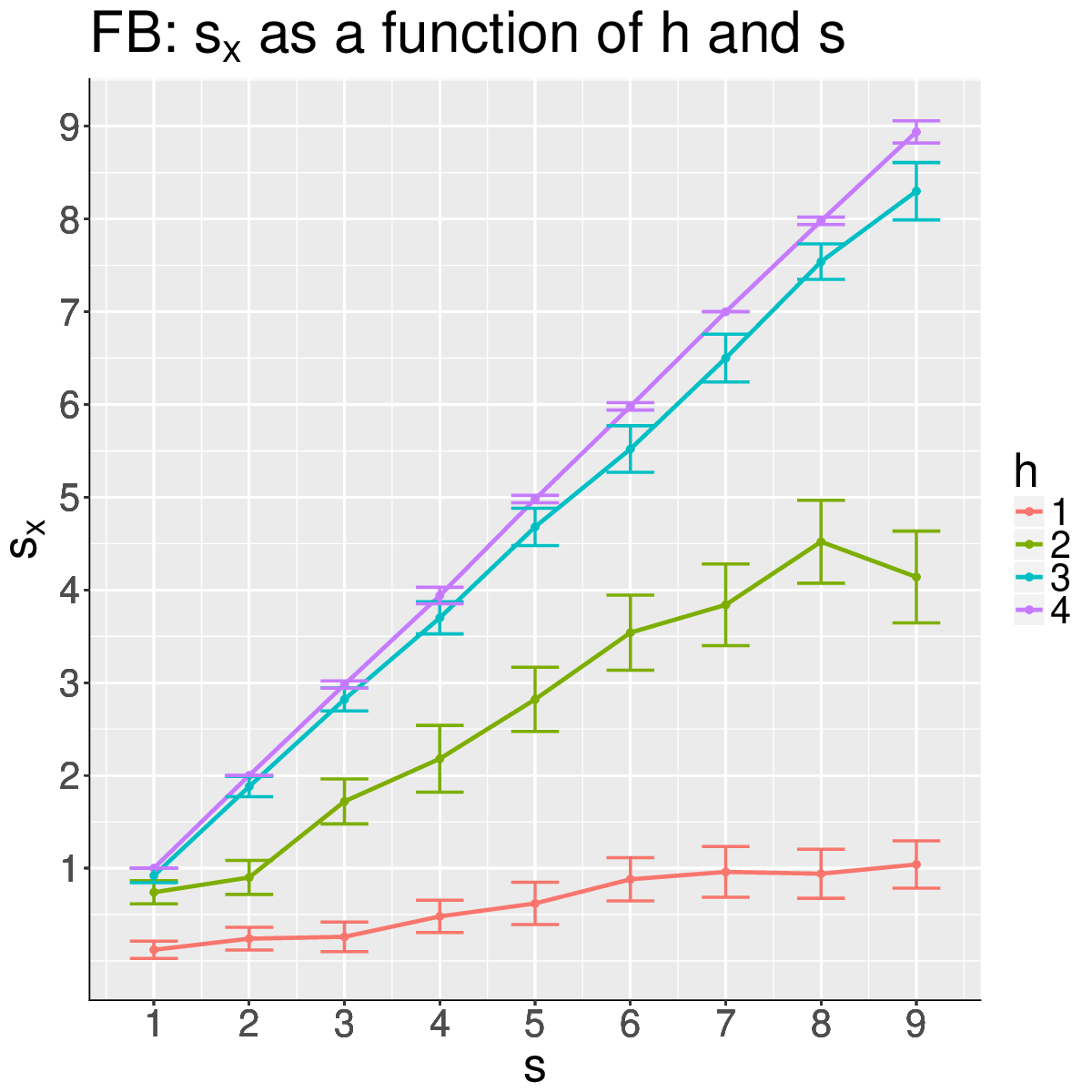}
    \caption{
            Facebook network \cite{mafoba15}.
    }
    \label{fig:fb_funhs}
\end{subfigure}
\caption{
    Average size of $S_x$ over 50 Monte Carlo simulations as a function
    of $(h,s)$; seed sets and vertex of interest selected randomly.
    As $h$ increases, more vertices in the graph are
    within an $h$-path of the randomly chosen VOI, and therefore more of
    the seed vertices are in $N_h(x)$. 
}
\label{fig:sxhvary}
\end{figure}

\begin{remark}{\it
Figure \ref{fig:sxhvary}\hspace{-5pt}  
demonstrates how $|S_x|$ depends on $|S|$ and
$h$ for graphs generated from a stochastic blockmodel (Figure
\ref{fig:sxhvary}\hspace{-5pt}a, model described in 
Definition \ref{def:sbm}) and for the Facebook network of
\cite{mafoba15} which we consider in detail in Section \ref{sec:hs}.
In both cases, the seed sets and
VOI are chosen uniformly at random. 
As expected, as $h$
increases, $|S_x|$ approaches $|S|$.
It is important
to keep in mind that increasing $h$ also increases $\ell$ and,
consequently, the sizes of $N_\ell(S_x)$ and $N_\ell(S_x')$, 
increasing computational complexity.
In both the simulated and Facebook examples, $h=2$ seems an
appropriate choice, and is the value we use for the networks in further
exploration (see Section \ref{sec:exp}).}
\end{remark}

\section{Simulations and real data experiments}
\label{sec:exp}
Note here that all necessary code and data needed to produce the figures
in this section can be found at \url{http://www.cis.jhu.edu/~parky/D3M/VNSGM/}.

We will measure the performance of \texttt{VNmatch} via rank$(x')$, the
expected rank of $x'$ in $\Phi_x$ when ties are broken uniformly at random.
Since the size of the set of candidate matches $C_x':=V_x'\setminus S_x'$ 
(seeds in $G'$ will {\it never} be matched to $x$ by \texttt{SoftSGM})
varies greatly in each experiment, we will 
compare across experiments by computing the normalized rank of $x'$ 
\begin{equation}
    \label{eqn:normrank}
    \tau(x') = \left(\frac{\text{rank}(x')-1}{|C_x'|-1}\right)\vee 0\in[0,1].
\end{equation}
Note that $\tau(x')=0$ (resp., $\tau(x')=0.5$ or $\tau(x')=1$) implies that 
the $\Phi_x(1)=x'$ (resp., $\Phi_x(|C_x'|/2)=x'$ or $\Phi_x(\alpha)=x'$ for $\alpha\geq |C_x'|$); i.e., the VOI was first, half-way down, or effectively last in the nomination list.
A low value of $\tau(x')$ corresponds to a low ranking of $x'$ in the
nomination list output from the \texttt{VNmatch} algorithm and
corresponds to a measure of how much time is saved (versus a uniformly
random search) by the end-user when searching through
the candidate set of vertices for the true match $x'$. 
We view a score of $\tau(x') = 5/100$ as better than a score of
$\tau(x') = 5/10$ since the amount of time saved by the end-user is
greater in the first case. 

\subsection{Simulation experiments}
\label{sec:sbmexp}

\begin{figure}
\centering
\begin{subfigure}[b]{0.4\textwidth}
    \includegraphics[width=\textwidth]{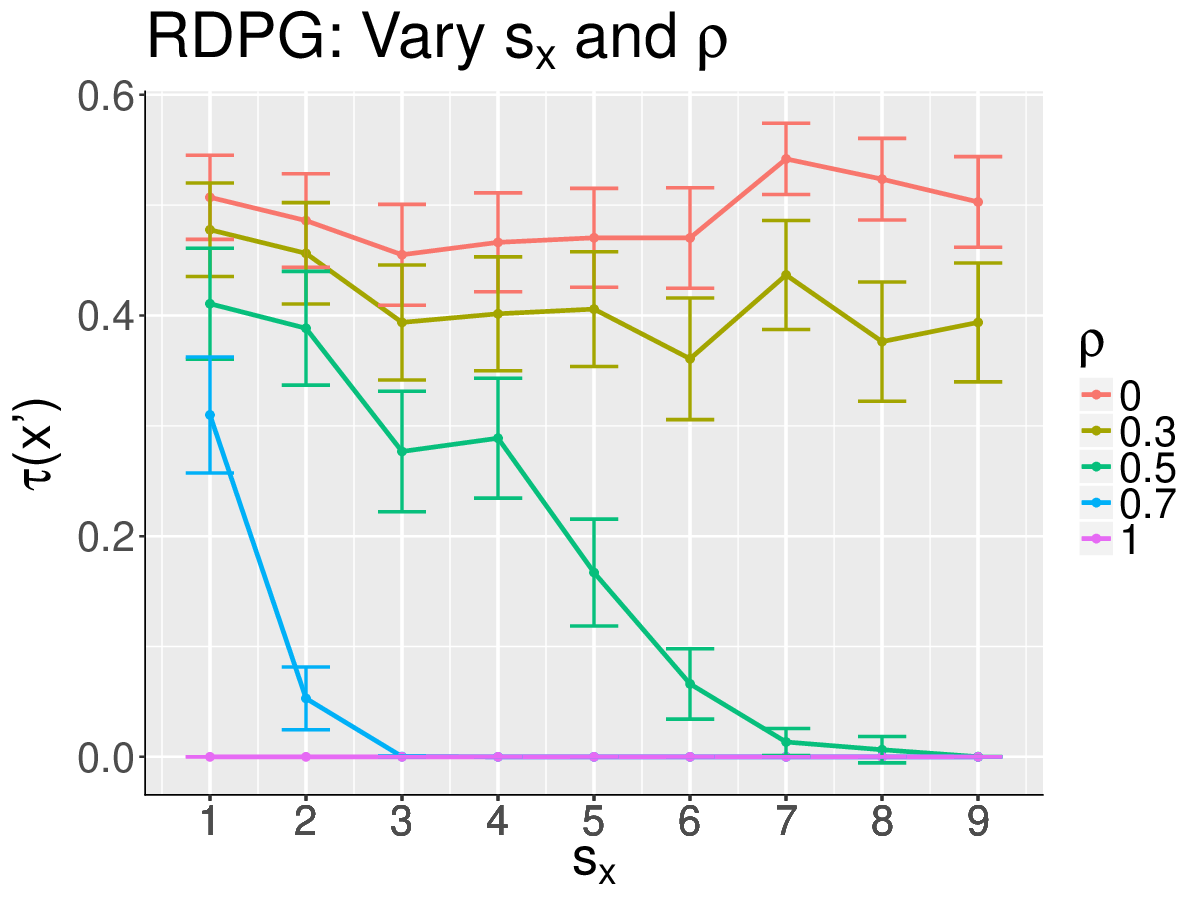}
    \caption{
            We plot $\tau(x')$ as a function of the number of seeds,
            $s_x$ for various $\rho$.
    }
    \label{fig:rdpg_seedvary}
\end{subfigure}
\quad\quad
\begin{subfigure}[b]{0.4\textwidth}
    \includegraphics[width=\textwidth]{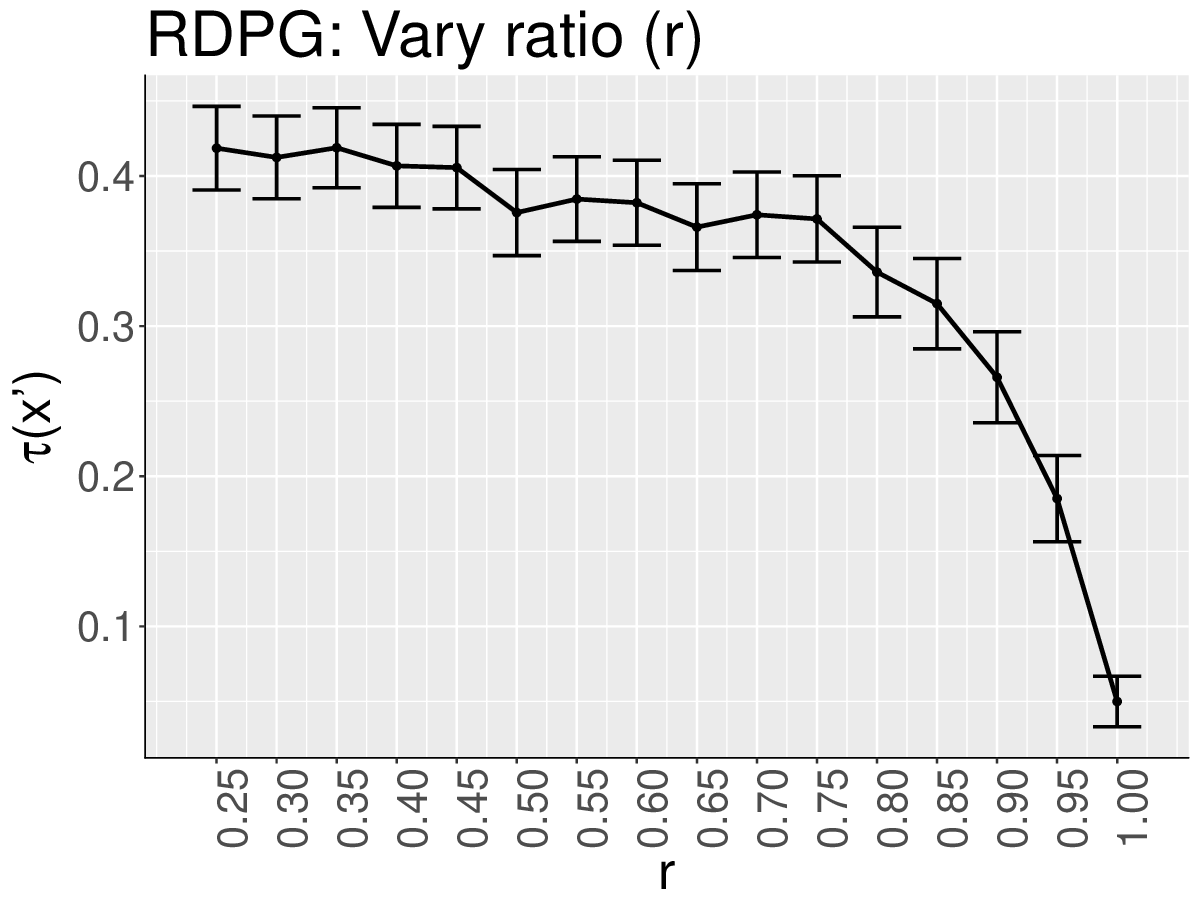}
    \caption{
            $\tau(x')$ as a
            function of $r$, the ratio of the vertex sizes of the two
            graphs, using $s_x=4$ and $\rho=0.6$, 
    }
    \label{fig:rdpg_ratiovary}
\end{subfigure}
\caption{
        For pairs of $300$-node graphs generated from a $\rho\mydash
        RDPG(X)$ model, we plot the average nomralized rank, $\tau(x')$,
        as a function of $s_x$, $\rho$, and $r$.
}
\label{fig:rdpg_srvary}
\end{figure}

We first explore the performance of Algorithm \ref{alg:vnsgm} in the 
$\rho$-RDPG setting, followed by the
$\rho$-SBM setting (see Section \ref{sec:rhocorr} for descriptions of
these models).
To wit, we first generate pairs of graphs from a $\rho\mydash RDPG(X)$,
where the latent positions of $X$ are uniformly chosen so that 
each row of $X$ is a unit vector and for any two rows of $X$, namely
$X_i$ and $X_j$, $X_i X_j^T \in (0,1)$. 
In Figure \ref{fig:rdpg_srvary}\hspace{-5pt} we explore how $\tau(x')$
is affected by the number of seeds used in the matching as compared
against various correlation values $\rho = 0,0.3,0.5,0.7,1$ 
(\ref{fig:rdpg_srvary}\hspace{-5pt}a)
and disparities in the sizes of the graphs to be matched when $\rho =
0.6$ (\ref{fig:rdpg_srvary}\hspace{-5pt}b). 

\begin{figure}
\centering
\begin{subfigure}[b]{0.4\textwidth}
    \includegraphics[width=\textwidth]{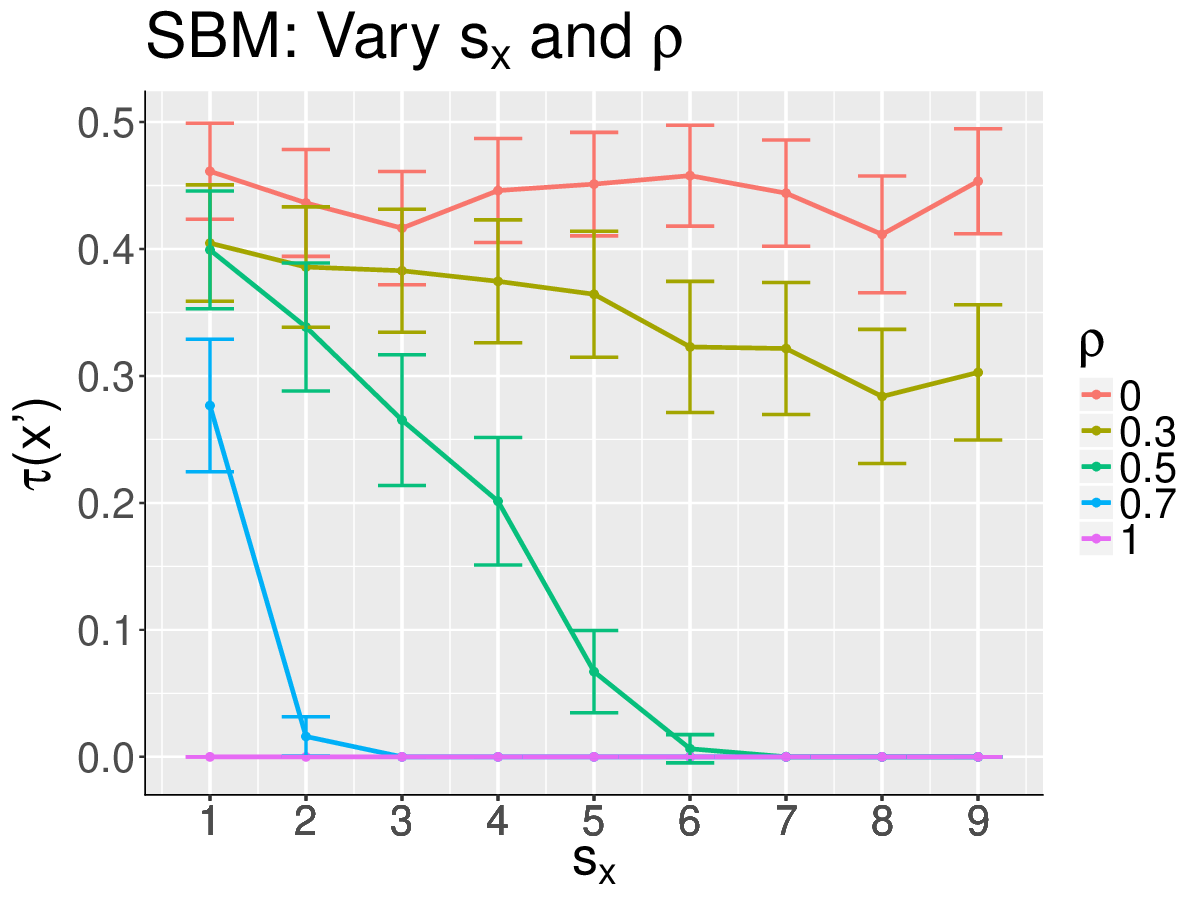}
    \caption{
            We plot $\tau(x')$ as a function of $s_x$ for various $\rho$.
    }
    \label{fig:seedvary}
\end{subfigure}
\quad\quad
\begin{subfigure}[b]{0.4\textwidth}
    \includegraphics[width=\textwidth]{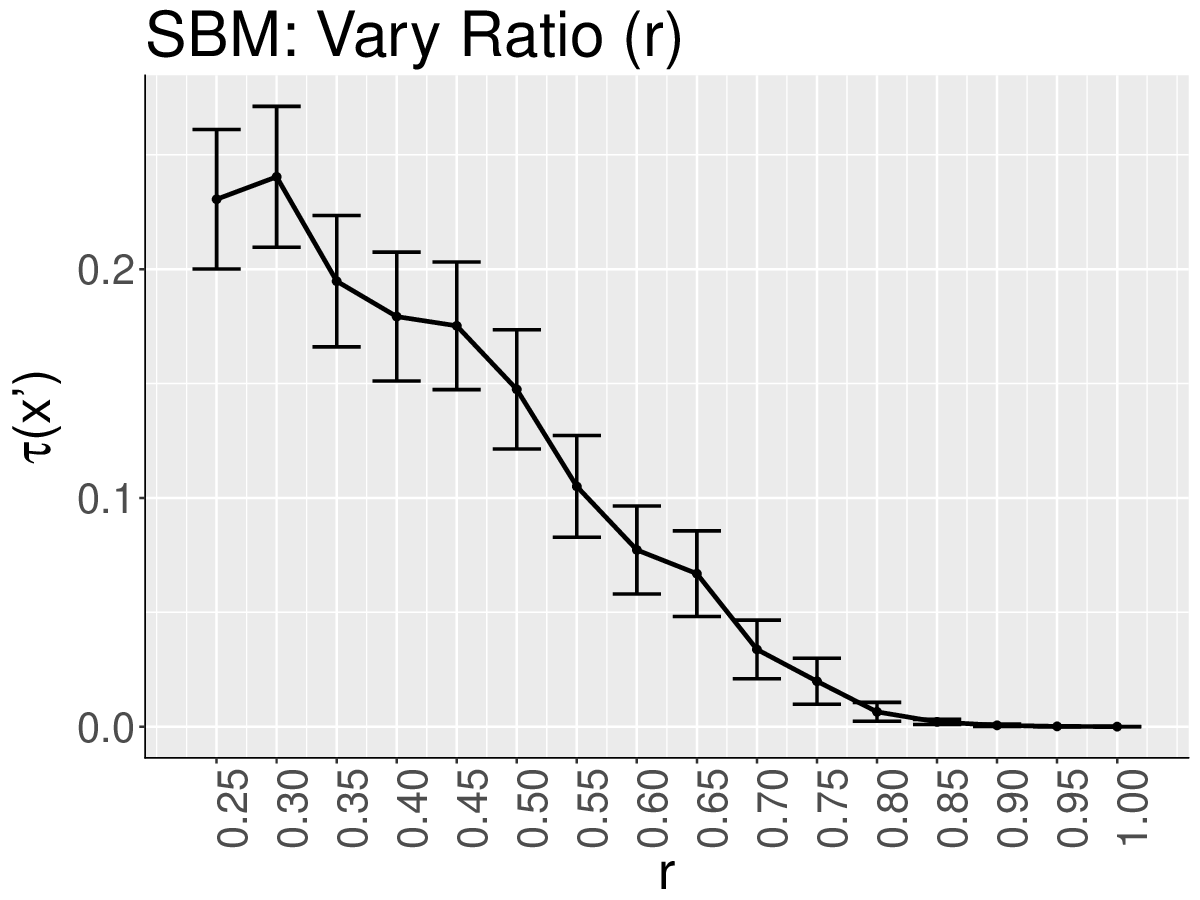}
    \caption{
            $\tau(x')$ as a function of $r$, the ratio of the vertex
            sizes of the two graphs, using $s_x=4$ and $\rho=0.6$, 
    }
    \label{fig:ratiovary}
\end{subfigure}
    \caption{
            For pairs of $300$-node graphs generated from a $\rho\mydash
            SBM(3,b,\Lambda)$, we the average normalized rank,
            $\tau(x')$, as a function of $s_x$, $\rho$, and $r$.
}
\label{fig:srvary}
\end{figure}

Next, we generate 
pairs of graphs from a $\rho\mydash SBM(3,b,\Lambda)$,
where $b$ is such that $1/3$ of the vertices are in each block and 
\begin{equation}
\label{eq:Lam}
    \Lambda = 
    \begin{bmatrix}
        0.7 & 0.3 & 0.4 \\
        0.3 & 0.7 & 0.3 \\
        0.4 & 0.3 & 0.7
    \end{bmatrix}.
\end{equation}
In Figure \ref{fig:srvary}\hspace{-5pt}, we explore how $\tau(x')$ is affected by 
the number of seeds used in the matching as compared against
correlation values $\rho = 0,0.3,0.5,0.7,1$
(\ref{fig:srvary}\hspace{-5pt}a) 
and disparities in the sizes of the graphs to be matched when $\rho =
0.6$ (\ref{fig:srvary}\hspace{-5pt}b). 

In order to explore how the number of seeds used in matching, $s_x$, 
affects the location of the VOI in the nomination list,
in both the RDPG and SBM setting,
we vary $s_x$ from 1 to 9, and run 100 Monte Carlo replicates using
\texttt{VNmatch}, with both the VOI and the 
seeds chosen uniformly at random in each Monte Carlo replicate. 
In Figures \ref{fig:rdpg_srvary}\hspace{-5pt}a and
\ref{fig:srvary}\hspace{-5pt}a, 
we record the average normalized rank of the VOI in the
nomination list ($\pm 2$s.e.) for the RDPG and SBM settings,
respectively.
It is apparent that
for sufficiently correlated networks,
as the number of seeds increases, 
our proposed nomination scheme becomes more accurate; 
i.e., the location of the VOI in the
nomination list is closer to the top of the list.
For graphs with very low correlation, the uniformly poor performance can be attributed to both the lack of much common structure between $G_x$ and $G_x'$ and the failure of \texttt{SoftSGM} to tease out this common structure.  
Since both $G$ and $G'$ are dense networks, $N_\ell(S_x)$ and
$N_\ell(S_x')$ generally contained between 250 and 300 vertices each.
Thus, the proportion of shared vertices in $G_x$ and $G_x'$
is rather high for this example.

To explore how the normalized rank of the VOI  
is influenced by 
matching graphs which differ in size,
we next consider pairs of graphs on different sized vertex sets.
We will set the number of vertices in the smaller graph, $G'$, 
to be $|V'|=r|V|=300r$,
for $r = 0.25, 0.30, \ldots, 1$. 
Let $H' = G'$ and suppose there exists an induced subgraph $H$ of
$G$ so that $(H,G') \sim 0.6 \mydash RDPG(X)$ for Figure
\ref{fig:rdpg_srvary}\hspace{-5pt}b and $(H,G') \sim 0.6 \mydash
SBM(3,b,\Lambda)$ for Figure \ref{fig:srvary}\hspace{-5pt}b.
For each $r$, 
we plot the average $\tau(x')$ ($\pm2$s.e.) over 100 Monte Carlo replicates
for fixed $s_x=4$.
As can be seen, under this model when the original networks $G$ and $G'$
have a large discrepancy between the sizes of their vertex sets there is
less accuracy in the \texttt{VNmatch} algorithm. 
Furthermore, the more obvious community structure present in the SBM
setting contributes to better performance of the \texttt{VNmatch}
algorithm.  Although we are not matching graphs $G_x$ and $G_x'$ with
vertex size difference ratio $r$ at every iteration, since the
connectivity of the vertices is high, $G_x$ and $G_x'$ do not deviate
much from being the full graphs. 

Li and Campbell explore the effects of utilizing seeds in graph matching
problems in \cite{lica15}. 
They found that although a
small number of seeds can greatly increase the number of correctly
matched vertices, as the number of shared users decreases so
does the ability to find a good match. 
As might be expected, since 
the number of potential mismatches increases as the
number of shared users decreases, 
Figures \ref{fig:rdpg_srvary}\hspace{-5pt}b and 
\ref{fig:srvary}\hspace{-5pt}b 
are consistent with Li and Campbell's results.

\begin{figure}
\centering
\begin{subfigure}[b]{0.41\textwidth}
    \includegraphics[width=\textwidth,trim=1cm 0 0 0]{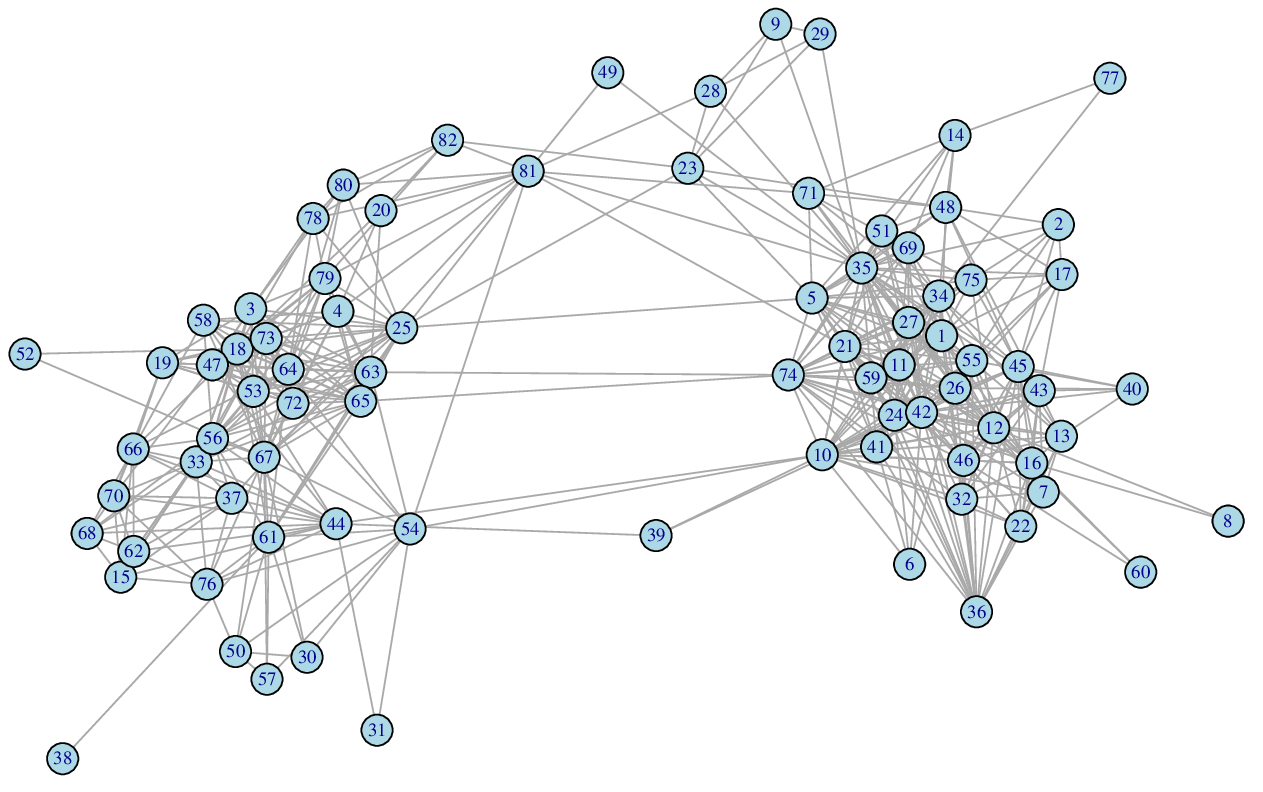}
    \caption{
            Facebook
    }
    \label{fig:hscfb}
\end{subfigure}
\quad\quad
\begin{subfigure}[b]{0.41\textwidth}
    \includegraphics[width=\textwidth,trim=1cm 0 0 0]{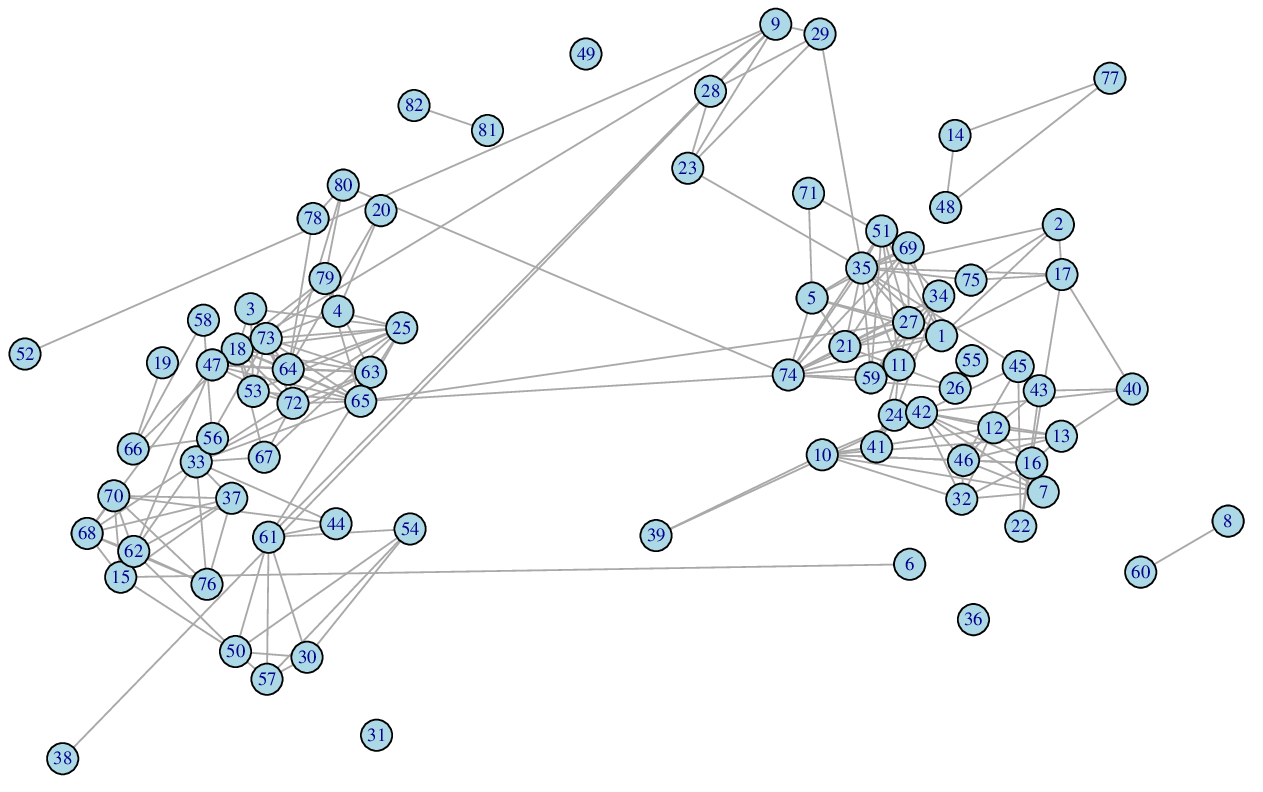}
    \caption{
            Survey
    }
    \label{fig:hscsurvey}
\end{subfigure}
\caption{
    The induced subgraphs for the High School Facebook and Survey
    networks generated by the shared vertices \cite{mafoba15}.
}
\label{fig:hscore}
\end{figure}

\subsection{Real data experiments}
\label{sec:realdata}

In this section, we explore two applications of \texttt{VNmatch} on real data.
Section \ref{sec:hs} explores a pair of high-school networks
obtained from \cite{mafoba15} 
in which the first graph is created 
based on student responses to a `who-knows-who' survey 
and the second is a Facebook friendship network involving some of the same students. 
In Section \ref{sec:it}, we consider Instagram and Twitter
networks having over-lapping vertex sets in which 
we would like to identify which Instagram profile corresponds to a
particular Twitter profile.
 
\begin{figure*}[t!]
\centering
\vspace{-10mm}
    \includegraphics[width=.85\textwidth]{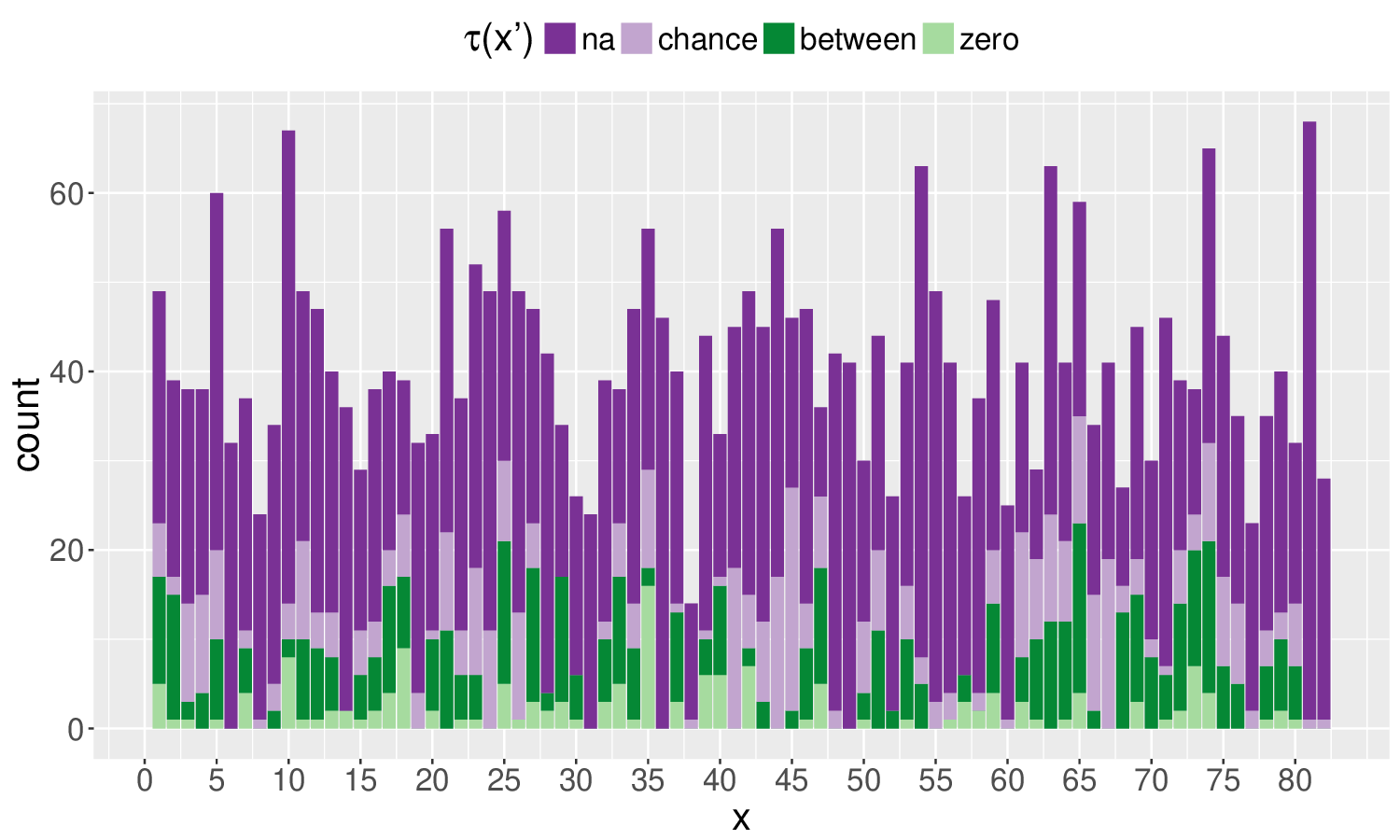} 
\caption{
    Consider each $x\in V$ as the VOI ($|V|=82$).
    For each $v \in N_2(x)$ 
    match the induced subgraphs of $H$ and $H'$ 
    generated by $N_2(v)$ and $N_2(v')$, considering $\{v\}$ and
    $\{v'\}$ to be seed-sets of size $1$.
    For each $x$, plot how often 
    $\tau(x') \in \{0, (0,0.5), [0.5,1], NA\}$ in light green, dark
    green, light purple, and dark purple, respectively 
    (colors here listed in order as they appear in the plot from bottom to top). 
    The height of the stack represents the total number of vertices
    in $N_2(x)$.
}
\label{fig:hsc_summary}
\end{figure*}

\subsubsection{Finding friends in high school networks} 
\label{sec:hs}

We consider two High School friendship networks on
over-lapping vertex sets published in \cite{mafoba15}. 
The first network, 
having 156 vertices, 
represents a Facebook network of profiles in which 
two vertices are adjacent if the pair of individuals were friends on Facebook.
The second network 
consists of 134 vertices, each representing a particular
student, and two vertices are adjacent if one of the students
reported that they are friends with the other student. 
There are 82 shared vertices across the two networks
for which we know the bijection between the two vertex sets, 
and the remaining vertices are known to have no such correspondence.
In the language of Section \ref{S:intro}, $\eta=156$, $\eta'=134$,
$n+s=81$, $m=74$, and $m'=52$.

Due to the large number of unshared vertices 
(nearly 40\% and 50\% for the Survey 
and Facebook networks, respectively), for illustrative purposes we 
perform our analysis of this data set by 
looking at the induced subgraphs generated by the 
shared vertices.
A brief glimpse into the effects of the unshared vertices can be found
in the supplemental material accompanying this article.
This step is purely for exploratory analysis and would not be
feasible in practice, as we would not have prior knowledge about which
vertices in the networks are shared as opposed to unshared.
At the same time, immediate success of \texttt{VNmatch} is still
not guaranteed since
the structure of the two graphs is very different, see Figure 
\ref{fig:hscore}\hspace{-5pt}.
Furthermore, we can see that there appears to be a 2-block structure for
each of the (shared) networks, although, if we were to model these
networks the block probability matrices for the two networks appears to
differ (unlike our simulation examples). 

\begin{figure}[t!]
\centering
\begin{subfigure}[b]{0.4\textwidth}
    \includegraphics[width=\textwidth,trim=1cm 0 0 0]{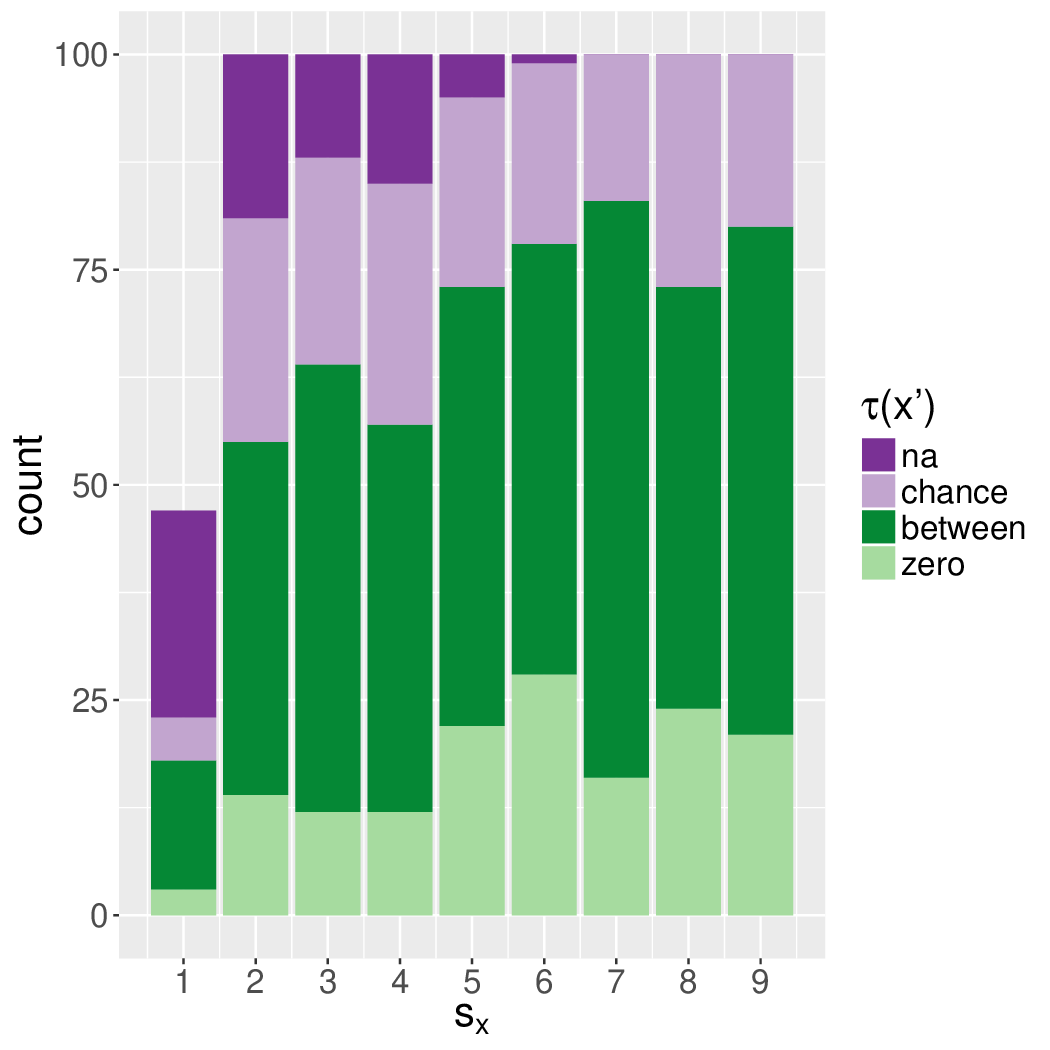} 
    \caption{
        As a function of $s_x$, plot
        how often $\tau(x') \in \{0, (0,0.5),[0.5,1],NA\}$ in
        light green, dark green, light purple, and dark purple, respectively
        (colors in order as they appear in plot from bottom to top).
    }
\end{subfigure}
\quad
\begin{subfigure}[b]{0.4\textwidth}
    \includegraphics[width=\textwidth,trim=1cm 0 0 0]{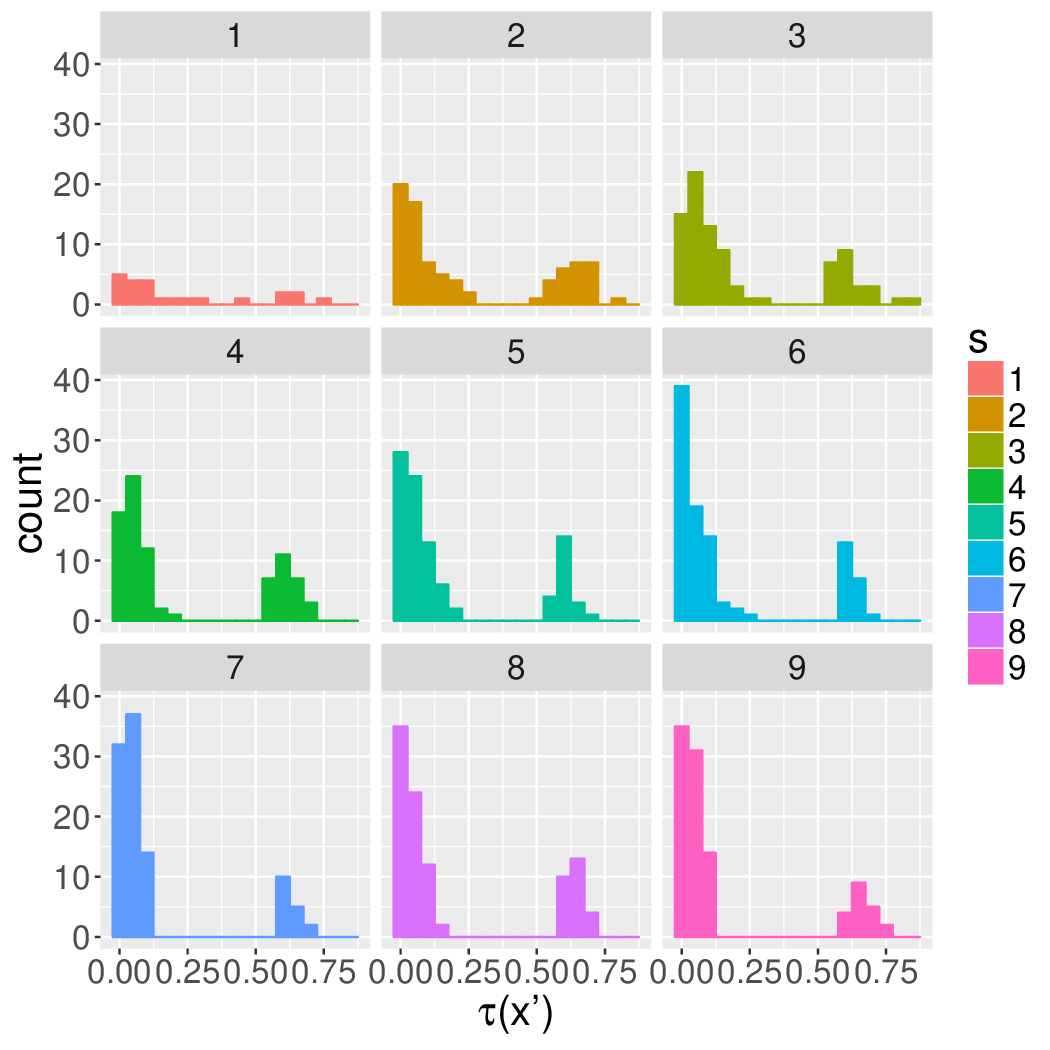}
    \caption{ 
        For each $s_x$ we show a bar-chart of $\tau(x')$ for each
        simulation in which $x' \in V_x'$ (i.e., $\tau(x') \ne NA$).
    }
\end{subfigure}
\caption{
    Using $x=v_{27}$ as the VOI, vary $s_x$ from $1$ to $9$ in \texttt{VNmatch}.  
    For $s_x>1$, uniformly at random generate $100$ seed sets from $N_2(x)$. 
    For $s_x=1$, consider all $47$ possible seed sets of size $1$.
}
\label{fig:net}
\end{figure}

We first explore how \texttt{VNmatch} performs when 
finding the VOI using a single seed. 
%
    Let $H$ and $H'$ denote, respectively, the induced subgraphs of the 
    High School Facebook and Friendship-Survey networks
    generated by the $82$ shared vertices.
    We run $82$ experiments, one for considering each $x \in V$ as the VOI, and for each VOI we consider using each $v \in N_2(x)$ as our single seed for \texttt{VNmatch}. 
    In Figure \ref{fig:hsc_summary}\hspace{-5pt}, for each $x$, we plot how often 
    $\tau(x') \in \{0, (0,0.5), [0.5,1], NA\}$ in light green, dark
    green, light purple, and dark purple, respectively 
    (colors listed in order as they appear in Figure
    \ref{fig:hsc_summary}\hspace{-5pt} from bottom to top): 
    When $\tau(x') = 0$, the
    true match $x'$ is at the top of the nomination list -- this is the
    best case possible; when $\tau(x') \in (0, 0.5)$, $x'$ is somewhere
    between the top of the nomination list and half-way down (i.e.
    better than chance, but not first); when $\tau(x') \in [0.5,1]$ the nomination list from \texttt{VNmatch} is worse than a uniformly random nomination list; and finally $\tau(x') = NA$ means that 
    $x' \not \in V_x'$ and our algorithm cannot hope to nominate the correct vertex. 
    The height of the stack represents the total number of vertices
    in $N_2(x)$.
    While beyond the scope of this work, this figure points 
    to the impact of seed-selection as
    well-chosen seeds can be the difference between perfect algorithmic
    performance and performance worse than chance.
    Note also that for vertices 
    6, 31, 36, and 49, $x' \not \in V_x'$ for all 
    $v\in N_2(x)$, so, matching the two neighborhoods for these
    vertices would never be successful for $\ell=h=2$.

We next consider the effects of increasing $s_x$.
For simplicity, we present our findings while
considering vertex 27 to be the VOI.
Vertex 27 shows moderately good performance using 1 seed in Figure
\ref{fig:hsc_summary}\hspace{-5pt}, although not the best.
We expect \texttt{VNmatch} to work equally well on any other vertex with
similar (or better) performance to vertex 27 as noted in Figure 
\ref{fig:hsc_summary}\hspace{-5pt}.

With vertex $x=v_{27}$ as the VOI in $G$, for each $s_x$ increasing from
$2$ to $9$ we uniformly at random generate $100$ seed sets from $N_2(x)$
and apply \texttt{VNmatch} to match $G_x$ and $G_x'$ using these seed
sets.  For $s_x=1$, rather than having 100 Monte Carlo
replicates, we consider only the $47$ possible seed sets of size $1$ in
$N_2(27)$.  Figure \ref{fig:net}\hspace{-5pt}, 
displays $\tau(x')$ as a function of $s_x$, with Figure
\ref{fig:net}\hspace{-5pt}a 
showing the general performance of $\tau(x')$ with
respect to $s_x$ and Figure \ref{fig:net}\hspace{-5pt}b 
displaying a frequency histogram (conditioned on $\tau(x')\neq NA$) of
$\tau(x')$ for each $s_x \in \{1,\ldots, 9\}$.

\begin{figure}[t!]
\centering
\begin{subfigure}[b]{0.42\textwidth}
    \includegraphics[width=\textwidth]{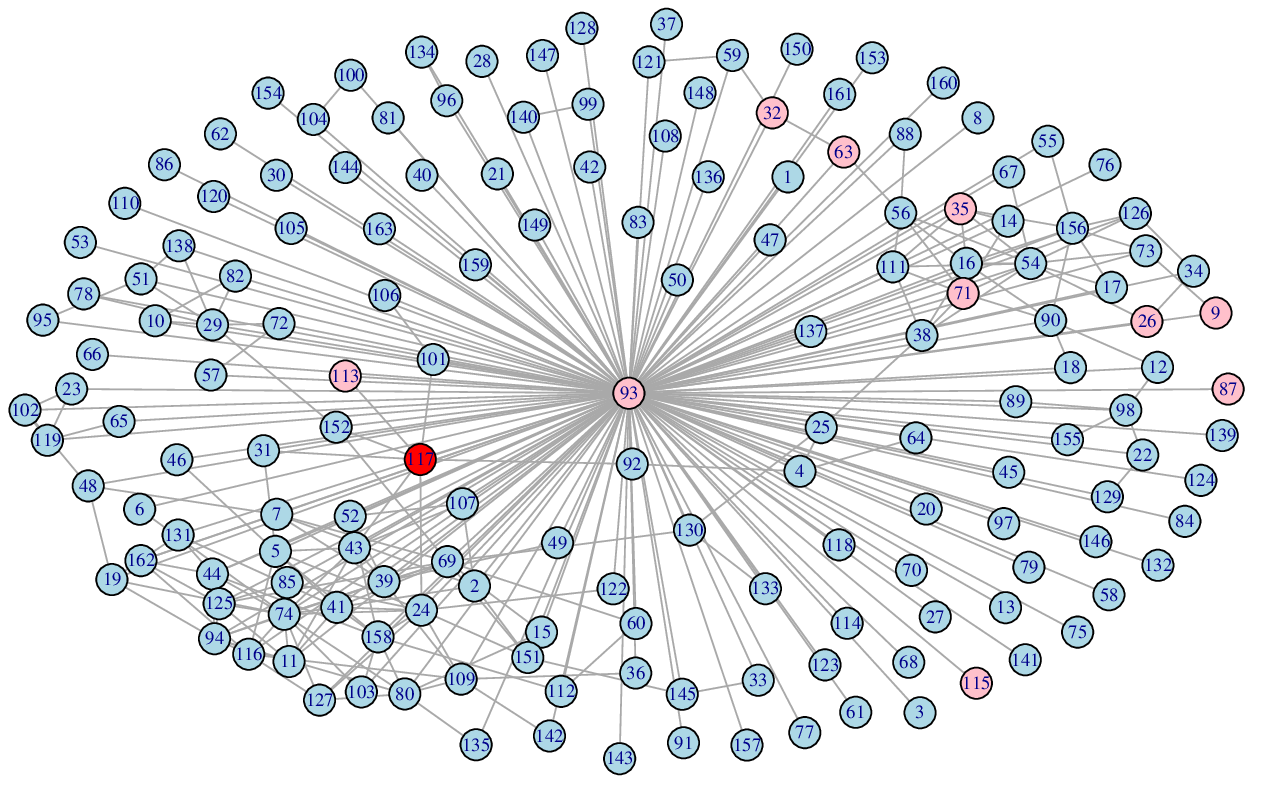}
    \caption{Twitter}
    \label{fig:twit}
\end{subfigure}
\quad\quad\quad\quad
\begin{subfigure}[b]{0.24\textwidth}
    \includegraphics[width=\textwidth]{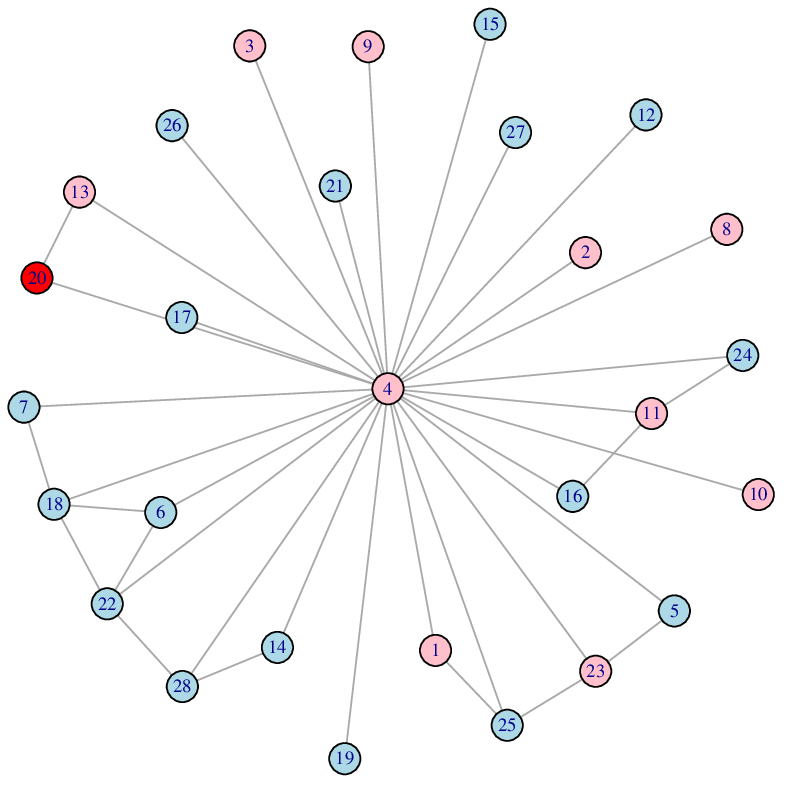}
    \caption{Instagram}
    \label{fig:insta}
\end{subfigure}
\caption{Graphs of a particular friend of the VOI for both Twitter and
    Instagram; VOI in red and seeds in pink.  }
\label{fig:it}
\end{figure}

\subsubsection{Finding Friends on Instagram from Twitter} 
\label{sec:it}

\begin{figure}[t!]
\centering
\includegraphics[width=0.45\textwidth]{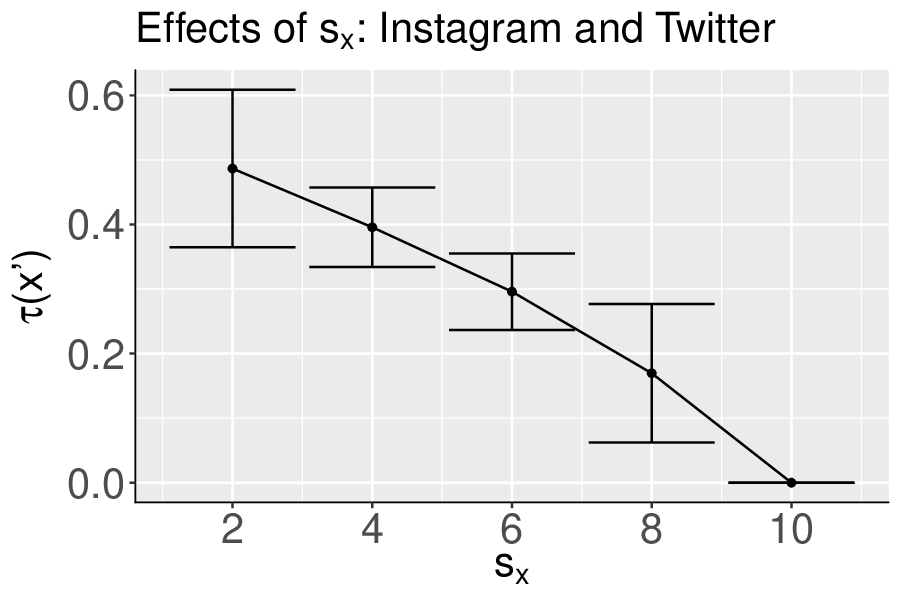}
\caption{ 
    Plot the average value of $\tau(x)$ (Equation
    \ref{eqn:normrank}), along with a confidence bound, 
    for $s_x=2,4,6,8,10$, always using the center vertex
    ($8$) as a seed.
}
\label{fig:itnoms_seeds}
\end{figure}

We next consider nominating across two publicly available social network
datasets, one derived from Twitter and one derived from Instagram, where
there is an edge between two vertices if one vertex is following the
other vertex in the respective social network.
We consider a single vertex present on both the Twitter and Instagram
networks and construct the two-hop neighborhoods of this vertex in each
network, yielding a 163 vertex Twitter graph (Figure
\ref{fig:it}\hspace{-5pt}a) and
a 28 vertex Instagram graph (Figure \ref{fig:it}\hspace{-5pt}b). 
After identifying a VOI in each network, a simple metadata analysis of vertex features yields 10 potential seeds.
In Figure \ref{fig:itnoms_seeds}\hspace{-5pt}, we plot
the average value of $\tau(x')$ ($\pm2$s.e.) when using a seed set of size $s_x=2,4,6,8,10$.
To avoid pathologies arising from $x'\not\in V_x'$, we use vertex 8 as a seed in each experiment.
As there are few seeds here, we average $\tau$ over all possible sets of seeds of size $s_x$ in each example.

There are a few takeaways from this figure.  
First note that as the number of seeds increases, the performance of
\texttt{VNmatch} increases significantly (i.e., the rank of $x'$ in
$\Phi_x$ is closer to the top).
In fact, 
we find that there are two vertices (including the
central vertex in both graphs) whose presence in the seed sets are
crucial in that if they are in the 
seed set then $\tau(x')=0$ every time, and if not then $\tau(x')>0.5$.
Thus, the improvement upon $\tau(x')$ in Figure
\ref{fig:itnoms_seeds}\hspace{-5pt}
is due to the increased proportion of seed sets which contain the two
crucial seeds for identifying the true match.
Furthermore, these are the only two seeds which are adjacent to the
vertex of interest.
This indicates that in the future it may be beneficial to focus on
what vertex-properties impact seed-usefulness in terms of assistance with
matchability.
Also note that these graphs are quite local---the full Twitter and
Instagram networks would have $>>10^7$ vertices---yet our algorithm
still performed quite well only considering $\approx 10^2$ vertices.  
Indeed, by whittling the networks down into local neighborhoods, we are
able to leverage the rich local signal present across networks without
the computational burden induced by working with the full, often
massive, networks themselves.

\section{Conclusions}
\label{sec:conc}

In this paper, we introduce an across-graph vertex nomination scheme
based on local neighborhood alignment  
for identifying a vertex of interest.
Our algorithm operates locally within much larger networks, and can scale to be implemented in the very large networks ubiquitous in this age of big data.
We demonstrated the efficacy of 
our principled methodology on both simulated and real data
networks, including an application to networks from Twitter and
Instagram.

In this paper we have focused on finding a corresponding vertex
in a second network to the VOI in the first network with
a notion of correspondence in our real-data examples meaning that two
nodes across the networks represent the same individual. 
Another application of this algorithm would be finding vertices, 
either across
two networks or across two subnetworks of one larger network, that have
\textit{similar structural role} across the two networks. 
Since the resulting nomination list of the \texttt{VNmatch} 
algorithm already outputs nodes in an ordering that is based 
on which vertices in a localized version of the second network 
have similar localized structural role to the 
VOI in the first network, this 
extension
follows immediately.

In the future, we would like to theoretically and empirically 
explore the impacts of network correlation and errors on \texttt{VNmatch} for
various random graph models.
We are also actively seeking to understand 
the effects of different types of seeds and what makes a ``good'' seed.
The impact of unshared vertices and their connections on the performance
of the \texttt{VNmatch} algorithm is still an open area of
investigation.
Applying \texttt{VNmatch} to multiple VOI could be done either
iteratively or simultaneously.  Other questions to explore include the
addition of attributes and how to apply \texttt{VNmatch} simultaneously
across multiple (more than $2$) networks.

\section*{Acknowledgments}
This work is partially supported by the XDATA and D3M and
SIMPLEX programs of the Defense Advanced Research Projects Agency (DARPA),
the Acheson J. Duncan Fund for the Advancement of Research in Statistics 
(Awards 16-20 and 16-23 and 18-3), and by EPSRC grant no. EP/K032208/1.
This material is also based on research sponsored by the Air Force
Research Laboratory and DARPA, under agreement number FA8750-18-2-0035.
The U.S. Government is authorized to reproduce and distribute reprints
for Governmental purposes notwithstanding any copyright notation
thereon. The views and conclusions contained herein are those of the
authors and should not be interpreted as necessarily representing the
official policies or endorsements, either expressed or implied, of the
Air Force Research Laboratory and DARPA, or the U.S. Government.
The authors would also like to thank the Isaac
Newton Institute for Mathematical Sciences, Cambridge, UK, for support and
hospitality during the program for Theoretical Foundations for Statistical
Network Analysis where a portion of work on this paper was undertaken.
The authors would also like to thank Jason Matterer for his helpful
comments and suggestions.

\subsection*{Author contributions}

Conceived and designed the methodology: HP VL CP. Performed the
experiments: HP YP. Analyzed the data: HP YP. Wrote the paper: HP VL CP.


\subsection*{Financial disclosure}

None reported.

\subsection*{Conflict of interest}

The authors declare no potential conflict of interests.

\section*{Supporting information}

Relevant code and data for all simulations and experiments can be found at 
\url{http://www.cis.jhu.edu/~parky/D3M/VNSGM/}.

The high school friendship and Facebook network data was originally
published in \cite{mafoba15} as data sets three and four, respectively.
A detailed description of the data sets can also be found at
\url{http://www.sociopatterns.org/datasets/high-school-contact-and-friendship-networks/}.

\bibliography{vnsgmbib}%

\section*{Author Biography}

\begin{biography}{\includegraphics[width=60pt,height=60pt,clip,keepaspectratio]{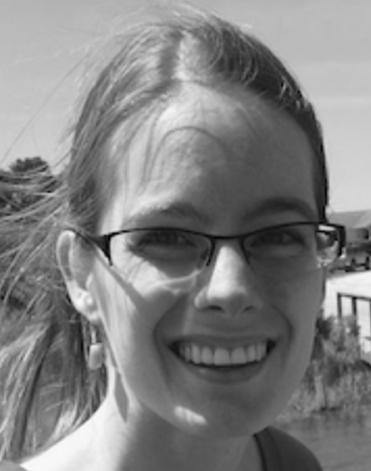}}{{Heather G. Patsolic}
    received the BS degree in mathematics from Wingate University, in
    2012, and the MA degree in mathematics from Wake Forest University,
    in 2014. She is currently a doctoral candidate in the Applied
    Mathematics and Statistics Department at Johns Hopkins University
    (JHU).
    Her research interests include statistical inference for
    high-dimensional and graph data, model selection, and pattern
    recognition.}
\end{biography}

\begin{biography}{\includegraphics[width=60pt,height=70pt,clip,keepaspectratio]{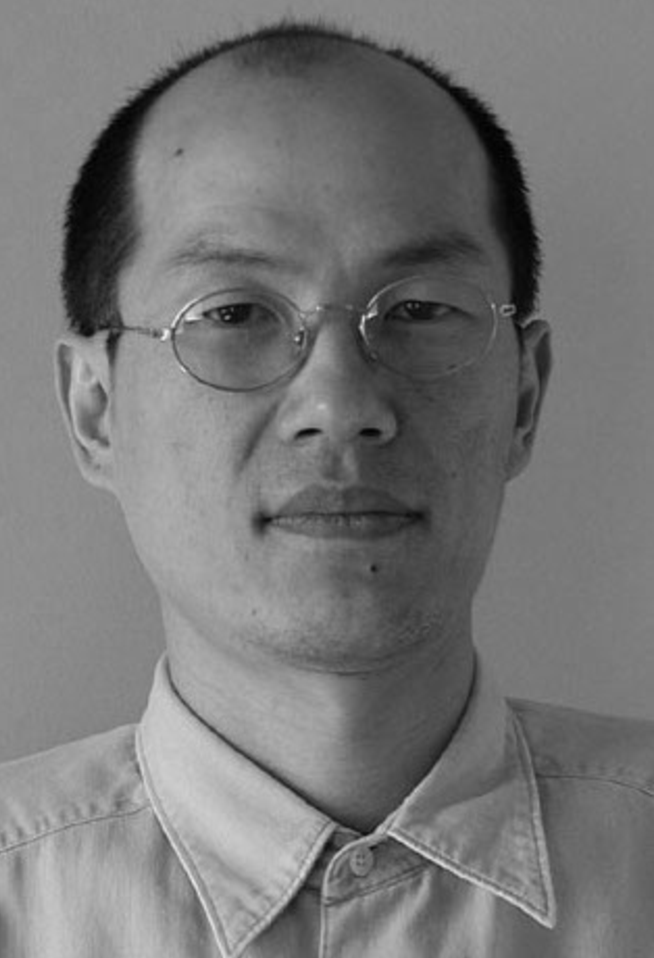}}{{Youngser Park}
    received the BE degree in electrical engineering from Inha
    University, Seoul, Korea, in 1985, and the MS and PhD degrees in
    computer science from George Washington University, in 1991 and
    2011, respectively. From 1998 to 2000, he worked in the Johns
    Hopkins Medical Institutes as a senior research engineer. From 2003
    until 2011, he worked as a senior research analyst, and has been an
    associate research scientist since $2011$ in the Center for Imaging
    Science, Johns Hopkins University (JHU). At JHU, he holds joint
    appointments in the Institute for Computational Medicine and the
    Human Language Technology Center of Excellence. His current research
    interests include clustering algorithms, pattern classification, and
    data mining for high-dimensional and graph data.}
\end{biography}

\begin{biography}{\includegraphics[width=60pt,height=70pt,clip,keepaspectratio]{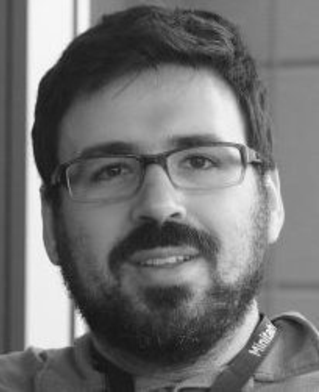}}{{Vince Lyzinski}
    received the BSc degree in mathematics from the University of Notre
    Dame in 2006, the MSc degree in mathematics from Johns Hopkins
    University (JHU) in 2007, the MSE degree in applied mathematics and
    statistics from JHU in 2011, and the PhD degree in applied mathematics
    and statistics from JHU, in 2013. From 2013 to 2014, he was a
    postdoctoral fellow in the Applied Mathematics and Statistics (AMS)
    Department, JHU, and from 2014 to 2017, he was a senior
    research scientist in the JHU Human Language Technology Center of
    Excellence and an assistant research professor in the AMS
    Department, JHU.  From 2017 to 2019 he was an assistant professor in
    the Department of Mathematics and Statistics at the University of
    Massachusetts, Amherst. Since 2019 he has been an assistant
    professor in the Department of Mathematics at the University of
    Maryland, College Park. His research interests include graph
    matching, statistical inference on random graphs, pattern
    recognition, dimensionality reduction, stochastic processes, and
    high-dimensional data analysis.}
\end{biography}

\begin{biography}{\includegraphics[width=60pt,height=70pt,clip,keepaspectratio]{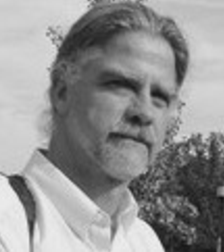}}{{Carey E. Priebe}
    received the BS degree in mathematics from Purdue University in
    1984, the MS degree in computer science from San Diego State
    University, in 1988, and the PhD degree in information technology
    (computational statistics) from George Mason University, in 1993.
    From 1985 to 1994, he worked as a mathematician and scientist in the
    US Navy research and development laboratory system. Since 1994, he
    has been a professor in the Department of Applied Mathematics and
    Statistics, Johns Hopkins University (JHU). At JHU, he holds joint
    appointments in the Department of Computer Science, Department of
    Electrical and Computer Engineering, Center for Imaging Science,
    Human Language Technology Center of Excellence, and Whitaker
    Biomedical Engineering Institute. His research interests include
    computational statistics, kernel and mixture estimates, statistical
    pattern recognition, statistical image analysis, dimensionality
    reduction, model selection, and statistical inference for
    high-dimensional and graph data. He is a lifetime member of the IMS,
    an elected member of the ISI, and a fellow of the ASA.}
\end{biography}

\end{document}